\documentclass[a4paper, 10pt, conference]{ieeeconf}      
\usepackage{FG2026}
\let\labelindent\relax
\usepackage{graphicx}
\usepackage{multirow}
\usepackage{arydshln}
\usepackage{graphicx}
\usepackage{subcaption}
\usepackage{tikz}
\usepackage{enumitem}
\usepackage{amsmath}
\usepackage{amssymb}
\usepackage{booktabs}
\newcommand*\circled[1]{\tikz[baseline=(char.base)]{
            \node[shape=circle,draw,inner sep=1pt] (char) {#1};}}
\FGfinalcopy 

\IEEEoverridecommandlockouts                              
\def\FGPaperID{467} 

\title{\LARGE \bf
ViT-FREE: Efficient Face Recognition via Early Exiting and Synthetic Adaptation
}

\author{\parbox{16cm}{\centering
    {\normalsize Tahar Chettaoui$^{1,2}$,Guray Ozgur$^{1,2}$, Eduarda Caldeira$^{1,2}$,  Naser Damer$^{1,2}$, and  Fadi Boutros$^{1}$}\\
    {\normalsize
    $^1$ Fraunhofer Institute for Computer Graphics Research IGD, Germany, $^2$ Department of Computer Science, TU Darmstadt, Germany}}
    \thanks{This research work has been funded by the German Federal Ministry of Education and Research and the Hessen State Ministry for Higher Education, Research and the Arts within their joint support of the National Research Center for Applied Cybersecurity ATHENE.}
}

\usepackage{fancyhdr}

\begin{document}

\ifFGfinal
\thispagestyle{empty}
\pagestyle{empty}
\else
\author{Anonymous FG2026 submission\\ Paper ID \FGPaperID \\}
\pagestyle{plain}
\fi
\maketitle
\thispagestyle{fancy}
\begin{abstract}
Vision Transformers (ViTs) have gained significant attention in computer vision and shown strong potential for face recognition (FR). However, their high computational cost makes deployment on resource-constrained devices challenging, motivating the need for methods that balance efficiency and accuracy. In this work, we investigate early exiting in pretrained ViTs as a simple yet effective training-free strategy for efficient FR inference. Leveraging the uniform feature dimensionality across transformer encoder blocks, we introduce \textbf{ViT-FREE}, a multi-exit framework that enables face verification directly from intermediate representations without modifying or retraining the backbone model, and thus, reducing inference cost. Our approach is motivated by the intrinsic behavior of ViTs, where transformer blocks iteratively refine representations within a shared representation space. Empirically, we show that patch embeddings and attention maps evolve progressively across depth, exhibiting high similarity between consecutive ViT blocks and increasing alignment with the final representation. This indicates gradual feature refinement and attention convergence, suggesting that intermediate layers already provide stable and discriminative representations suitable for early exiting. Through extensive experiments on multiple FR benchmarks, we systematically analyze the accuracy-efficiency trade-off across exit depths. Our results demonstrate that later exits achieve a highly favorable balance, with exiting at layer 10 yielding up to a 20\% speedup while incurring only a $\sim$1.5\% drop in verification performance on benchmarks such as IJB-C. Also, we propose \textbf{ViT-FREE$_{FT}$}, a lightweight exit-specific fine-tuning strategy that adapts only the projection layers using a small synthetic dataset while keeping the transformer backbone frozen. This approach improves the performance of shallow exits while preserving the efficiency benefits and leaving deeper exits largely unaffected. Code and pretrained models will be publicly released.
\end{abstract}

\section{Introduction}
Vision Transformers (ViTs) \cite{dosovitskiy2021imageworth16x16words} have recently emerged as a powerful architecture in computer vision, achieving strong performance across a wide range of tasks, including Face Recognition (FR) \cite{DBLP:conf/cvpr/KimS0JL24, DBLP:journals/ivc/ChettaouiDB25,DBLP:conf/iccv/DanLXD0XS23}. Despite their strong recognition capabilities, ViTs remain computationally demanding, posing significant challenges for deployment in real-world applications such as mobile and edge devices with limited resources. Their high inference cost, largely driven by the global self-attention mechanism \cite{DBLP:conf/cvpr/Guo0WT00X22}, makes efficiency a key concern. Consequently, improving the efficiency of ViT-based FR systems without sacrificing accuracy has become a critical research problem.

To address high computational demands of ViT, prior work has proposed various efficiency-oriented strategies, including token pruning and merging \cite{DBLP:conf/cvpr/Tang00XGXT22, DBLP:conf/cvpr/ChangWLWZ0S23, DBLP:journals/corr/abs-2202-07800, DBLP:conf/iclr/BolyaFDZFH23}, architectural redesigns \cite{DBLP:journals/corr/abs-2112-07658, DBLP:conf/cvpr/LiuPZ00Y23, DBLP:conf/iccv/GrahamETSJJD21, DBLP:conf/iclr/MehtaR22}, and dynamic inference techniques \cite{DBLP:conf/icpr/Teerapittayanon16, DBLP:conf/mm/XuHSHLLS23, DBLP:journals/nn/BakhtiarniaZI22, DBLP:conf/biosig/NixonRCLT23, DBLP:conf/nips/WolczykWBPTST21}. While these approaches have demonstrated effectiveness in improving ViT efficiency for general computer vision tasks, they do not explicitly address or study the specific requirements and challenges of FR.

\begin{figure}[!t]
  \centering
   \includegraphics[width=1\linewidth]{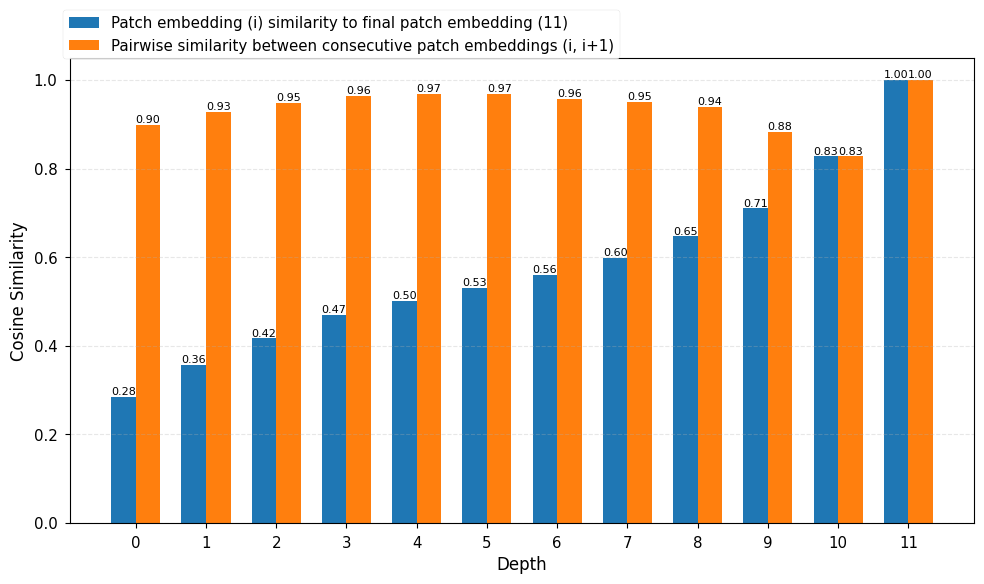}
   \caption{Cosine similarity of patch embeddings $\mathbf{h}_i$ at different depths (0–11) in a ViT, averaged over the LFW benchmark. The blue bars show the similarity between each intermediate patch embedding $\mathbf{h}_i$ and the final patch embedding $\mathbf{h}_{11}$, while the orange bars indicate the pairwise similarity between consecutive patch embeddings $\mathbf{h}_i$ and $\mathbf{h}_{i+1}$. The $x$-axis represents the depth of the ViT, and the $y$-axis shows the cosine similarity. High similarity between consecutive layers (orange bars) indicates small incremental refinement of features, while the steadily increasing similarity to the final patch embedding (blue bars) shows that each layer progressively converges toward the final representation.}
   \label{fig:plot_feature_refinement}
\vspace{-6mm}
\end{figure}

Early exiting has recently emerged as a promising dynamic inference strategy that enables termination at intermediate network depths \cite{DBLP:conf/bmvc/BakhtiarniaZI21}, reducing inference cost. In ViTs, this paradigm is particularly well-suited due to their uniform architectural design across layers, which allows intermediate representations to be directly reused for prediction without structural modifications. As detailed in Section \ref{sec:method}, this suitability is supported by three key properties: (1) \textbf{dimensional compatibility}, where all transformer blocks produce representations in a consistent $\mathbb{R}^{N \times D}$ space, enabling the shared projection head $g(\cdot)$ to operate at any depth \cite{DBLP:journals/corr/VaswaniSPUJGKP17}, (2) \textbf{progressive feature refinement}, where each layer incrementally refines token embeddings within the same feature space, leading to steadily increasing similarity to deeper representations and the final embedding, and (3) \textbf{attention evolution}, where attention maps gradually transition from diverse, less structured patterns in early layers to more stable and semantically consistent patterns in deeper layers, indicating convergence of spatial focus across depth. Together, these properties indicate that ViTs are particularly well-suited to early-exiting strategies.
Despite these favorable properties, most existing ViT-based FR approaches rely exclusively on the final-layer representation, potentially overlooking useful information available at intermediate depths and incurring unnecessary computational cost. Furthermore, the behavior of ViT representations across depth in the context of FR and its implications for early exiting remains underexplored.

In this work, we address this gap by systematically investigating  early exiting in pretrained ViTs for FR. We introduce \textbf{ViT-FREE}, a simple yet effective framework that enables face verification directly from intermediate representations without modifying or retraining the backbone network. Our study provides both empirical and analytical insights into how ViT representations evolve across depth and how this evolution can be exploited to achieve an optimal trade-off between accuracy and efficiency. In addition, we propose a lightweight fine-tuning strategy, \textbf{ViT-FREE$_{FT}$}, which adapts exit-specific projection layers using a small synthetic dataset to further enhance the performance of early exits.

Our contributions are summarized as  follows:
\begin{itemize}
    \item We conduct a detailed analysis of depth-wise feature refinement in ViTs for FR, showing that consecutive layers exhibit high similarity and that intermediate representations progressively converge toward the final embedding, naturally supporting early exiting.
    \item We perform a comprehensive evaluation of all exits of the considered ViT backbone across multiple FR benchmarks, quantifying both recognition performance and computational cost, and providing a detailed characterization of the accuracy-efficiency trade-off at each depth.
    \item We propose ViT-FREE$_{FT}$. This lightweight exit-specific fine-tuning strategy adapts only the projection layers using a small synthetic dataset while keeping the transformer backbone frozen, significantly improving the performance of shallow exits without full retraining.
    \item We provide qualitative and quantitative analyses of attention map evolution across depth, offering interpretable insights into how facial representations and spatial attention progressively stabilize throughout the network.
\end{itemize}

\begin{figure}[!t]
  \centering
   \includegraphics[width=0.9\linewidth]{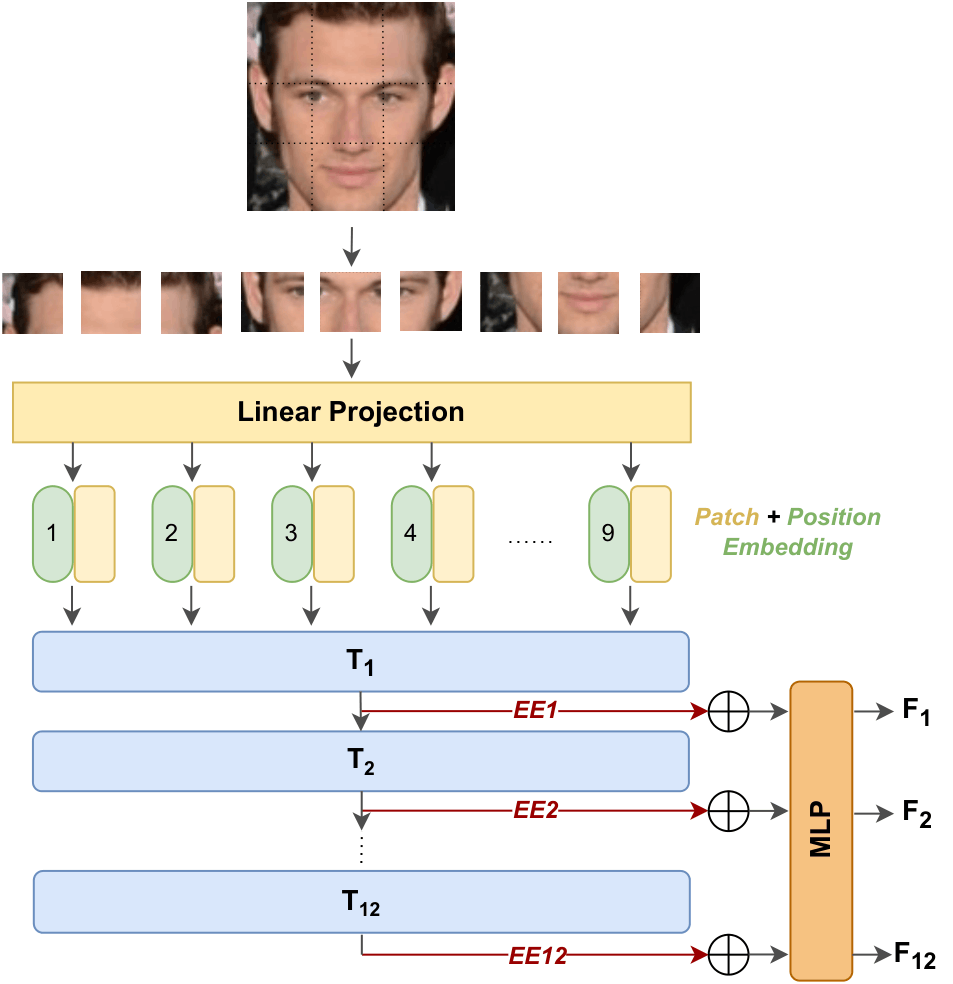}
   \caption{ViT-FREE inference pipeline. Intermediate patch embeddings from each transformer block are projected to feature representations, enabling predictions at different depths and allowing a flexible trade-off between computational efficiency and recognition performance. The symbol $\oplus$ denotes the concatenation operator.} 
   \label{fig:vit}
   \vspace{-6mm}
\end{figure}

\section{Related Work}
\subsection{Vision Transformers for Face Recognition}

ViTs, originally proposed for image classification \cite{dosovitskiy2021imageworth16x16words}, have recently emerged as a compelling alternative to CNNs, offering greater flexibility and the ability to model global contextual relationships, encouraging a range of solutions \cite{DBLP:conf/bmvc/SunT22, DBLP:conf/isie/KhanSEG23, DBLP:conf/cvpr/KimS0JL24, DBLP:conf/iccv/DanLXD0XS23, DBLP:journals/tcsv/QinWDWCHD24, DBLP:journals/corr/abs-2209-08930, DBLP:journals/corr/abs-2103-14803, DBLP:journals/ivc/ChettaouiDB25, DBLP:conf/bmvc/IslamZM22} focused on their use in FR. Sun \cite{DBLP:conf/bmvc/SunT22} proposed a part-based pipeline where a lightweight CNN predicts facial landmarks, and local patches around them are fed to a ViT for part-aware FR. FRoundation \cite{DBLP:journals/ivc/ChettaouiDB25} adapted CLIP \cite{radford2021learningtransferablevisualmodels} and DINOv2 \cite{DBLP:journals/tmlr/OquabDMVSKFHMEA24} foundation models for FR using Low-Rank Adaptation (LoRA) \cite{DBLP:conf/iclr/HuSWALWWC22}, demonstrating the potential of leveraging the inherent generalizability of foundation models in low-data scenarios. ARTriViT \cite{DBLP:conf/isie/KhanSEG23} is a triplet loss-based Siamese network with a ViT backbone that analyzes pairs of face images and uses similarity indexes for FR. To make ViT more resilient to scale, translation, and pose variations, KP-RPE \cite{DBLP:conf/cvpr/KimS0JL24} extends Relative Position Encoding by assigning pixel importance based on positions relative to facial keypoints, enhancing spatial relationship preservation under affine transformations. TransFace \cite{DBLP:conf/iccv/DanLXD0XS23} addresses overfitting and training instability when training ViTs on large-scale datasets such as MS-Celeb-1M \cite{guo2016ms} and Glint360K \cite{DBLP:conf/iccvw/AnZGXZFWQZZF21}, introducing Dominant Patch Amplitude Perturbation to enhance generalization and Entropy-based Hard Sample Mining to improve training efficiency. Despite these advances, most existing approaches rely exclusively on final-layer representations, potentially overlooking informative signals available at intermediate depths. 
Despite these advances, most existing ViT-based approaches for FR rely exclusively on final-layer representations, potentially overlooking informative signals available at intermediate depths. Moreover, the computational cost of ViTs remains a major limitation for practical FR deployment \cite{DBLP:conf/cvpr/Guo0WT00X22, DBLP:conf/cvpr/KimS0JL24}. Notably, prior work primarily focuses on improving recognition accuracy or robustness, while the problem of efficiency in ViT-based FR remains underexplored.

\subsection{Early Exits in ViTs}

Early exiting \cite{DBLP:conf/icpr/Teerapittayanon16} has emerged as an effective strategy, allowing predictions to be produced from intermediate layers without traversing the full model, thereby reducing computation. ViTs process images through a sequence of self-attention blocks, each progressively refining the representation of the input \cite{dosovitskiy2021imageworth16x16words}. Prior work has shown that intermediate representations in ViTs capture complementary levels of abstraction: early layers focus on local low-level patterns (e.g., edges and textures), similar to the initial layers in CNNs; middle layers capture object parts and spatial relations; and deeper layers encode high-level semantic concepts useful for classification \cite{dosovitskiy2021imageworth16x16words, raghu2021vision}. This hierarchical refinement suggests that many inputs may become linearly separable well before reaching the final layer \cite{eesurvey3}. Early-exit mechanisms leverage this property by allowing inference to terminate at intermediate network depths, skipping the computation of remaining layers, and such dynamic inference approaches aim to reduce latency and energy consumption while preserving accuracy \cite{eesurvey3, eesurvey4}. ViTs are particularly well-suited to early exits because all transformer blocks output tokens of the same dimensionality, enabling the straightforward insertion of auxiliary classifiers after any block without the need for complex feature adaptation \cite{DBLP:conf/bmvc/BakhtiarniaZI21}. Moreover, since each block refines the same set of tokens, exiting early often provides usable predictions with minimal architectural modification. 

Several techniques have been proposed to further improve early-exit performance for ViTs, including specialized branch architectures with transformer encoders or convolutional heads \cite{DBLP:journals/nn/BakhtiarniaZI22}, Single-Layer ViT (SL-ViT) mechanisms that fuse local and global patch information \cite{DBLP:journals/nn/BakhtiarniaZI22}, heterogeneous exit strategies like LGViT that employ different head designs for shallow versus deep layers \cite{DBLP:conf/mm/XuHSHLLS23}, self-distillation training approaches where branches mimic the final classifier's behavior \cite{distillation_multiexit}, and entropy-based dynamic routing policies that determine the optimal exit point based on prediction confidence \cite{BERxiT}. These methods enhance the discriminative power of intermediate classifiers and mitigate the accuracy gap between early and final exits. 
Nixon et al. \cite{DBLP:conf/biosig/NixonRCLT23,DBLP:journals/ejivp/NixonRCLT25} investigated early exiting in the context of closed-set face identification and analyzed how intermediate exits interact with bias-related effects in the model. 


Despite the progress in both ViT-based FR and early-exit strategies, the intersection of these two directions remains underexplored. Existing FR approaches predominantly rely on final-layer representations, while early-exit methods are largely developed for general vision tasks and do not study or evaluate early-exit FR. 
In this work, we bridge this gap by systematically investigating early exiting in ViTs for FR. We provide a comprehensive analysis of the accuracy-efficiency trade-off across all exit depths and demonstrate that intermediate representations can be effectively leveraged for face verification. 

\begin{figure}[!t]
  \centering
   \includegraphics[width=0.9\linewidth]{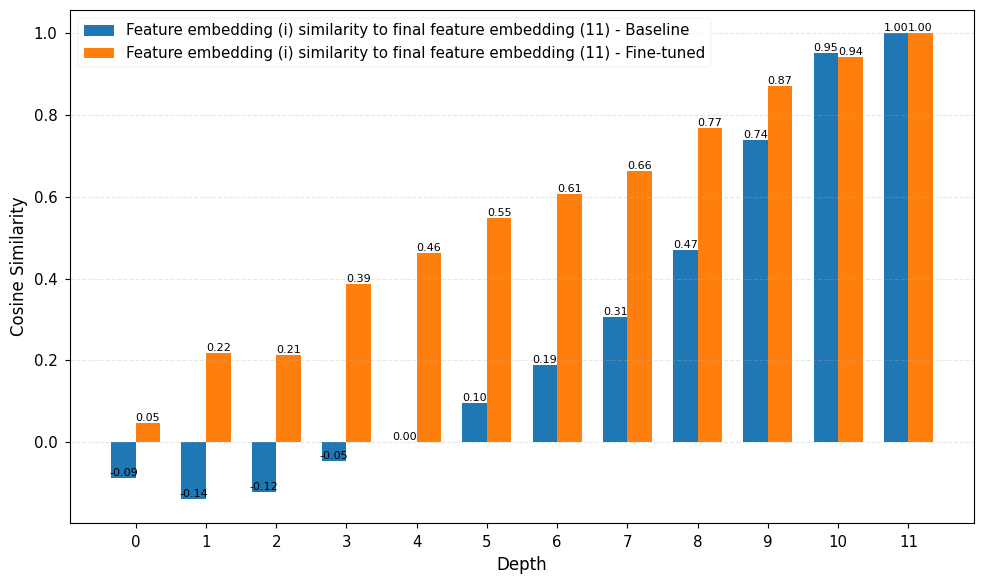}
   \caption{Cosine similarity between intermediate feature embeddings at each ViT depth (0–10) and the final embedding (depth 11) for both a baseline ViT-FREE and its fine-tuned variant ViT-FREE$_{FT}$, as defined in Section \ref{sec:experimental_setup}. ViT-FREE$_{FT}$ consistently achieves higher or comparable similarity to the final feature embedding across all depths, demonstrating the effectiveness of fine-tuning in accelerating the refinement of feature representations throughout the network.}
   \label{fig:plot_feature_embedding_similarity} 
   \vspace{-6mm}
\end{figure}

\section{Methodology} \label{sec:method}

We propose \textbf{ViT-FREE}, an early-exit framework for efficient FR based on ViTs. The key idea is to exploit the uniform feature dimensionality across transformer encoder blocks to enable predictions from intermediate representations without modifying the backbone. This enables dynamic inference with a controllable trade-off between computational efficiency and recognition performance.
An overview of ViT-FREE is shown in Figure \ref{fig:vit}. By attaching exits at multiple depths, the model can produce embeddings at different stages of computation, allowing early termination during inference without retraining or architectural changes.

\subsection{ViT (Preliminary)}

Given an input facial image \( \mathbf{X} \in \mathbb{R}^{H \times W \times C} \), it is divided into \( N = \frac{HW}{P^2} \) non-overlapping patches of size \( P \times P \). Each patch is flattened and linearly projected into a $D$-dimensional embedding space, resulting in:
\begin{equation}
\mathbf{Z}_0 \in \mathbb{R}^{N \times D}.
\end{equation}

To encode spatial information, learnable positional embeddings \( \mathbf{E}_{\text{pos}} \in \mathbb{R}^{N \times D} \) are added:
\begin{equation}
\mathbf{Z}_0 = \mathbf{Z}_0 + \mathbf{E}_{\text{pos}}.
\end{equation}

The transformer encoder consists of $L$ stacked blocks:
\begin{equation}
\mathbf{Z}_i = \mathcal{T}_i(\mathbf{Z}_{i-1}), \quad i = 1, \dots, L,
\end{equation}
where \( \mathcal{T}_i(\cdot) \) denotes a standard transformer encoder block composed of multi-head self-attention and Multi-Layer Perceptron (MLP) layers with residual connections.
To obtain the final representation, all patch embeddings are concatenated and projected into the FR embedding space. Let
\begin{equation}
\mathbf{Z}_L = [\mathbf{z}_L^1, \mathbf{z}_L^2, \dots, \mathbf{z}_L^N], \quad \mathbf{z}_L^k \in \mathbb{R}^{D}.
\end{equation}
The global representation is constructed as:
\begin{equation}
\mathbf{h}_L = [\mathbf{z}_L^1 \;\|\; \mathbf{z}_L^2 \;\|\; \dots \;\|\; \mathbf{z}_L^N] \in \mathbb{R}^{ND},
\end{equation}
where \( \|\) denotes concatenation.
The final embedding is obtained via a projection function:
\begin{equation}
\label{eq:projection}
\mathbf{F}_L = g(\mathbf{h}_L),
\end{equation}
where \( g: \mathbb{R}^{ND} \rightarrow \mathbb{R}^{D} \) is a learnable linear mapping.
The model is trained using a margin-based softmax loss (e.g., CosFace), which enforces inter-class separability and improves discriminative power.


\begin{figure}[!t]
  \centering
   \includegraphics[width=1\linewidth]{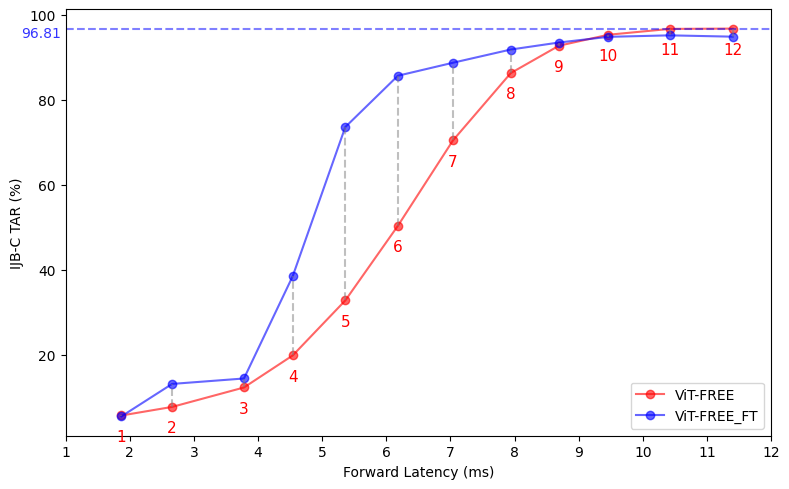}
   \caption{Effect of ViT early exits at different depths (1–12) on forward latency and FR performance. The x-axis represents the model’s forward latency in milliseconds, and the y-axis shows the IJB-C TAR measured at FAR=$10^{-4}$. Each point corresponds to a specific early-exit depth, illustrating the trade-off between computational latency and FR performance. The red curve represents the baseline ViT-FREE, while the blue curve corresponds to the fine-tuned variant ViT-FREE$_{FT}$. Fine-tuning substantially improves the performance of shallow exits, while both models converge to similar performance at deeper layers.}
   \label{fig:ijbc_latency}
   \vspace{-4mm}
\end{figure}

\subsection{ViT-FREE: Early Exit Formulation}

We implement early exits by extracting intermediate representations from each transformer block $l \in \{1, 2, \ldots, L\}$ without introducing additional parameters. The key insight underlying our early exit approach stems from a fundamental architectural difference between ViTs and CNNs. While CNNs progressively reduce spatial dimensions while increasing channel depth (e.g., from $224 \times 224 \times 3$ input to $112 \times 112 \times 64$, then $56 \times 56 \times 128$), ViTs maintain constant token dimensionality $D$ throughout all $L$ transformer layers. This architectural property enables three critical advantages for early exit implementation: 
(1) \textbf{Dimensional compatibility}: Every transformer block outputs representations in $\mathbb{R}^{N \times D}$ format, allowing direct application of pre-trained projection function $g(.)$ to any intermediate layer without architectural modifications \cite{DBLP:journals/corr/VaswaniSPUJGKP17}. 
(2) \textbf{Feature refinement}: Each transformer block refines embeddings within the same $D$-dimensional space \cite{dosovitskiy2021imageworth16x16words}, enabling the extraction of intermediate representations from any block $l \in \{1, 2, \ldots, L\}$ to use them as feature representation. Figure \ref{fig:plot_feature_refinement} shows the cosine similarity between patch embeddings across depths, averaged over the LFW benchmark \cite{huang:inria-00321923}. We observe that consecutive layers have high cosine similarity, indicating gradual, consistent updates, while the similarity to the final patch embedding increases with depth, reflecting progressive refinement toward the final representation. This demonstrates that ViT blocks perform incremental, layer-wise refinement, with intermediate patch embeddings converging to the final patch embedding. 
(3) \textbf{Attention pattern evolution}: Each transformer block produces an attention map over the input tokens, reflecting which regions the model attends to at a given depth. Figure \ref{fig:plot_attention_map_similarity} shows the cosine similarity between attention maps across depths, averaged over the LFW benchmark \cite{huang:inria-00321923}. Early layers exhibit low similarity to the final attention map, indicating that the attention patterns have not yet converged. As depth increases, this similarity grows, reflecting progressive convergence. The pairwise similarity between consecutive maps follows the same trend, indicating that later blocks apply smaller updates. This suggests that attention maps undergo substantial changes in early layers and converge toward the final depth.

For each transformer block \( i \), we define:
\begin{equation}
\mathbf{Z}_i = [\mathbf{z}_i^1, \mathbf{z}_i^2, \dots, \mathbf{z}_i^N].
\end{equation}

Similar to the final layer, we construct an intermediate representation by concatenating all token embeddings:
\begin{equation}
\mathbf{h}_i = [\mathbf{z}_i^1 \;\|\; \mathbf{z}_i^2 \;\|\; \dots \;\|\; \mathbf{z}_i^N] \in \mathbb{R}^{ND},
\end{equation}

and compute the corresponding embedding using the same projection layer in Equation \ref{eq:projection} :
\begin{equation}
\mathbf{F}_i = g(\mathbf{h}_i).
\end{equation}

This allows inference to terminate at any depth \( i \leq L \), and thus reduces computational cost.

\subsection{Lightweight Exit-Specific Embedding Fine-Tuning}
While ViT-FREE enables efficient inference in a training-free manner, intermediate representations, particularly at shallow depths, are not explicitly optimized for optimal identity discrimination. To address this limitation, we introduce a lightweight fine-tuning strategy that adapts the embedding layer for each exit using a small synthetic dataset.

\paragraph{Synthetic Data.}
We employ a compact synthetic dataset \cite{boutros2026idperturb}. This enables efficient refinement of the embedding space without requiring access to large-scale labeled data.

\paragraph{Exit-Specific Projection.}
In the original ViT-FREE formulation, a shared projection function \( g(\cdot) \) is used across all exits. We instead introduce exit-specific projection functions:
\begin{equation}
\mathbf{F}_i = g_i(\mathbf{h}_i), \quad i = 1, \dots, L,
\end{equation}
where each \( g_i: \mathbb{R}^{ND} \rightarrow \mathbb{R}^{D} \).

\paragraph{Optimization.}
Each projection head \( g_i \) is optimized independently using a margin-based softmax loss (e.g., CosFace \cite{wang2018cosfacelargemargincosine}), while all transformer encoder blocks \( \{\mathcal{T}_i\}_{i=1}^{L} \) remain frozen. This preserves the pretrained backbone while adapting each exit to produce more discriminative embeddings.

\paragraph{Efficiency.}
The proposed extension introduces minimal overhead, as only lightweight projection layers are trained. The inference complexity remains unchanged, apart from selecting the corresponding exit-specific projection layer. This preserves the efficiency advantages of ViT-FREE while improving performance, particularly for early exits.

\begin{table*}[!t]
\centering
\resizebox{0.9\linewidth}{!}{%
\begin{tabular}{c|cccccc|cc|cc|c}
\multirow{2}{*}{\centering Method} 
& \multirow{2}{*}{\centering LFW} 
& \multirow{2}{*}{\centering CFP-FP} 
& \multirow{2}{*}{\centering AgeDB30} 
& \multirow{2}{*}{\centering CALFW} 
& \multirow{2}{*}{\centering CPLFW} 
& \multirow{2}{*}{\centering Avg.} 
& \multirow{2}{*}{\centering IJB-B} 
& \multirow{2}{*}{\centering IJB-C} 
& \multicolumn{2}{|c|}{Tinyface} 
& \multirow{2}{*}{\centering GFLOP $\downarrow$} \\ 
\cline{10-11}
& & & & & & & & & Rank-1 & Rank-5 & \\ \midrule
ViT-FREE12 & 99.77 & 98.73 & 98.05 & 96.05 & 94.00 & 97.32 & 95.43 & 96.81 & 68.37 & 72.16 & 11.50  \\
ViT-FREE11 & 99.78 & 98.54 & 97.92 & 96.08 & 94.10 & 97.29 & 95.41  & 96.74 & 68.96 & 72.37 & 10.55\\
ViT-FREE10 & 99.68 & 97.33  & 96.60 & 95.53& 92.35 & 96.30 &  93.57 & 95.36 & 66.52 & 71.46 & 9.60\\
ViT-FREE9 & 99.52 &  95.44 & 93.35 & 94.20& 91.00 & 94.70 & 90.67  & 92.77 & 63.41 & 68.83 & 8.65\\
ViT-FREE8 & 98.85 & 92.51  & 89.15 & 91.68&  88.00 & 92.04 &  83.15 & 86.33 & 60.94 & 66.95 & 7.71\\
ViT-FREE7 & 97.53 &  86.69 & 82.58 & 86.78& 81.03 & 86.92 & 65.37  & 70.49 & 57.38 & 64.14 & 6.76\\
ViT-FREE6 & 94.10 & 78.24  & 74.42 & 80.53& 71.47 & 79.75 &  45.17 & 50.34 & 52.90 & 60.01 & 5.81\\
ViT-FREE5 & 90.25 & 72.01  & 69.02 & 74.33& 64.55 & 74.03 &  27.85 &  32.74  & 47.51 & 56.25 & 4.86 \\
ViT-FREE4 & 84.42 & 68.20  & 63.30 & 67.97& 58.98 & 68.57 & 16.89  & 19.78  & 41.63 & 50.91 & 3.91 \\
ViT-FREE3 & 81.30 &  64.51 & 60.07 & 63.45& 57.20 &  65.31 &   10.31 & 12.23 & 35.89 & 45.63 & 2.96  \\
ViT-FREE2 & 77.77 & 64.44  & 57.52 & 61.22& 57.15 & 63.62 & 6.58  & 7.65 & 29.51 & 38.44 & 2.01\\
ViT-FREE1 & 75.58 & 62.47  &55.37  &58.57 & 56.63 & 61.72 & 4.90  & 5.63 & 22.80 & 31.25 & 1.06\\ 
\midrule 
\midrule
ViT-FREE12$_{FT}$ & 99.70 & 97.50 & 96.28 & 95.28 & 92.70 & 96.29 & 93.06 & 94.91 & 68.08 & 71.89 & 11.50 \\
ViT-FREE11$_{FT}$ &  99.68 & 97.60 & 96.73 & 95.32 & 92.83 & 96.43 & 93.63 & 95.22 & 68.99 & 72.29 & 10.55\\
ViT-FREE10$_{FT}$ &  99.65 & 97.70 & 96.53 & 95.20 & 92.67 & 96.35 & 93.21 & 94.87 & 69.23 & 72.77  & 9.60 \\
ViT-FREE9$_{FT}$ &  99.65 & 97.44 & 96.37 & 94.62 & 92.13 & 96.04 & 91.38 & 93.53 & 66.55 & 70.60 & 8.65\\
ViT-FREE8$_{FT}$ &  99.67 & 96.80 & 95.22 & 94.60 & 91.37 & 95.53 & 89.50 & 91.88 & 66.55 & 70.60 & 7.71\\
ViT-FREE7$_{FT}$ &  99.55 & 95.43 & 94.02 & 94.12 & 90.27 & 94.68 & 86.11 & 88.77 & 64.97 & 68.94 & 6.76\\
ViT-FREE6$_{FT}$ &  99.25 & 93.56 & 92.23 & 93.10 & 88.55 & 93.34 & 82.15 & 85.70 & 61.53 & 66.58 & 5.81\\
ViT-FREE5$_{FT}$ &  99.13 & 91.57 & 90.37 & 92.47 & 86.32 & 91.97 & 73.16 & 73.61 & 58.64 & 64.43 & 4.86\\
ViT-FREE4$_{FT}$ & 98.68 & 88.94 & 88.58 & 91.35 & 84.32 & 90.38 & 43.84 & 38.46 & 55.04 & 61.00 & 3.91\\
ViT-FREE3$_{FT}$ &  98.27 & 84.03 & 84.93 & 89.72 & 80.08 & 87.41 & 19.24 & 14.35 & 50.78 & 57.24 & 2.96 \\
ViT-FREE2$_{FT}$ & 96.85 & 74.39 & 81.18 & 86.87 & 72.82 & 82.42 & 17.14 & 13.07 & 46.19 & 52.76 & 2.01\\
ViT-FREE1$_{FT}$ & 93.35 & 63.37 & 74.85 & 81.92 & 62.98 & 75.29 & 6.76 & 5.43 & 37.45 & 43.75 & 1.06\\
\end{tabular}}
\caption{Verification performance of ViT-FREE early exits at different depths (1–12) on several FR benchmarks, as presented in Section \ref{sec:experimental_setup}, along with its variant ViT-FREE$_{FT}$, which is fine-tuned on a small synthetic dataset, as described in Section \ref{sec:experimental_setup}. Results on IJB-B and IJB-C are reported as TAR at FAR of $10^{-4}$. ViT-FREE 10-11 provides the optimal trade-off between accuracy and efficiency.} 
\label{tab:results_table}
\vspace{-3mm}
\end{table*}

\begin{table*}[!t]
\centering
\resizebox{0.60\linewidth}{!}{%
\begin{tabular}{c|ccccc}
Method & GFLOP & Fwd latency (ms) & Speed-up $\uparrow$ & \#Param (M) & $\Delta$ in Param (M) \\ \midrule
ViT-FREE12 & 11.50 & 11.40 & 1.00 $\times$ & 76.02 & \circled{1} \\
ViT-FREE11 & 10.55 & 10.42 & 1.09 $\times$ & 72.87 & \circled{1} - 3.15 \\
ViT-FREE10 & 9.60 & 9.46 & 1.20 $\times$ & 69.72 & \circled{1} - 6.30  \\
ViT-FREE9 & 8.65 & 8.69 & 1.32 $\times$ & 66.57 & \circled{1} - 9.45 \\
ViT-FREE8 & 7.71 & 7.94 & 1.44 $\times$ & 63.42 & \circled{1} - 12.60\\
ViT-FREE7 & 6.76 & 7.04 & 1.62 $\times$ & 60.27 & \circled{1} - 15.75\\
ViT-FREE6 & 5.81 & 6.18 & 1.84 $\times$ & 57.12 & \circled{1} - 18.90\\
ViT-FREE5 & 4.86 & 5.36 & 2.12 $\times$ & 53.97 & \circled{1} - 22.05 \\
ViT-FREE4 & 3.91 & 4.54 & 2.51 $\times$ & 50.82 & \circled{1} - 25.20 \\
ViT-FREE3 & 2.96 & 3.78 & 3.01 $\times$ & 47.67 & \circled{1} - 28.35 \\
ViT-FREE2 & 2.01 & 2.66 & 4.28 $\times$ & 44.51 & \circled{1} - 31.51\\
ViT-FREE1 & 1.06 & 1.87 & 6.09 $\times$ & 41.36 & \circled{1} - 34.66 \\
\end{tabular}}
\caption{Computational resources of ViT-FREE early exits at different depths (1–12), including FLOPs, inference latency, and parameter count, are reported to compare efficiency across different exit depths. We measure it as \cite{DBLP:conf/kdd/RasleyRRH20}. Notably, ViT-FREE and ViT-FREE$_{FT}$ share identical computational cost, as the fine-tuning process only adapts the final projection layers without introducing additional parameters or modifying the backbone architecture.} 
\label{tab:results_table_computation}
\vspace{-3mm}
\end{table*}


\begin{table}[t!]
\centering
\resizebox{1\linewidth}{!}{%
\begin{tabular}{c|cccccc|c}
\multirow{2}{*}{\centering Method} 
& \multirow{2}{*}{\centering LFW} 
& \multirow{2}{*}{\centering CFP-FP} 
& \multirow{2}{*}{\centering AgeDB30} 
& \multirow{2}{*}{\centering CALFW} 
& \multirow{2}{*}{\centering CPLFW} 
& \multirow{2}{*}{\centering Avg.} 
& \multirow{2}{*}{\centering GFLOP $\downarrow$} \\ 
& & & & & \\ \midrule
KPRPE-FREE24 & 99.83 & 99.01 & 97.66 & 96.00  & 95.40 & 97.58 & 25.04 \\
KPRPE-FREE22 & 99.80 & 98.97 & 97.47 &  95.98 & 95.40 & 97.52 & 22.99\\ 
KPRPE-FREE20 & 99.82 & 98.80 & 96.63 & 95.78 & 95.07 & 97.22 & 20.93\\
KPRPE-FREE18 & 99.65 & 98.05 & 94.85 & 94.79 & 94.06 & 96.28  & 18.87\\
KPRPE-FREE16 & 99.50 & 96.63 & 91.63 & 93.15 & 92.87 & 94.76 & 16.81\\
KPRPE-FREE14 & 98.92 & 94.48 & 87.25 &  90.25 & 90.20 & 92.22 & 14.76\\
KPRPE-FREE12 & 97.97 & 91.08 & 81.22 & 87.18 & 86.23 & 88.74 & 12.70 \\
KPRPE-FREE10 & 96.22 & 86.18 & 76.30 & 82.99 & 80.37 & 84.41  & 10.64\\
KPRPE-FREE8 & 93.25 & 79.27 & 70.13 & 76.82 & 72.05 & 78.30 & 8.59 \\
KPRPE-FREE6 & 87.90 & 71.13 & 63.62 & 69.53 & 64.38 & 71.31 & 6.53 \\
KPRPE-FREE4 & 80.25 & 64.61 & 58.68 & 63.33 & 56.70  & 64.71 & 4.47 \\
KPRPE-FREE2 & 73.85 & 61.13 & 54.87 & 58.82 & 55.03 & 60.74 & 2.41\\
\end{tabular}}
\caption{Verification performance of ViT KP-RPE \cite{DBLP:conf/cvpr/KimS0JL24} early exits at different depths (1–24) on several FR benchmarks, as presented in Section \ref{sec:experimental_setup}. As ViT-B KPRPE comprises 24 transformer layers, results are reported at every second depth.} 
\label{tab:results_table_kprpe}
\vspace{-7mm}
\end{table}

\vspace{-1mm}
\section{Experimental Setup}
\vspace{-1mm}
\label{sec:experimental_setup}
\subsection{Training Settings} 
We use the ViT-S architecture with an input image size of $112$×$112$ and a patch size of $9$. The model employs an embedding dimension of $512$, $8$ attention heads, and a depth of $12$ transformer layers. To train our ViT model,  we employ the MS1MV2 \cite{DBLP:conf/cvpr/DengGXZ19} dataset. MS1MV2 is a refined version of the MS-Celeb-1M dataset \cite{guo2016ms}, and comprises 5.8M images of 85K identities. Following \cite{DBLP:journals/ivc/ChettaouiDB25}, we employ the CosFace loss function \cite{wang2018cosfacelargemargincosine}. Optimization is carried out using the AdamW optimizer \cite{loshchilov2019decoupledweightdecayregularization} with a weight decay of $0.05$. The model is trained for $40$ epochs using a batch size of $1024$. We set the initial learning rate to $0.001$ and adopt a polynomial learning rate schedule \cite{DBLP:conf/tencon/MishraS19} with a warmup period of $4$ epochs. To improve generalization, we apply a set of data augmentation techniques, following \cite{DBLP:conf/cvpr/KimS0JL24}, including horizontal flipping, brightness, contrast adjustments, scaling, translation, RandAugment \cite{Randaugment_CVPR}, Gaussian blur, cutout, and rotations up to $20^\circ$. Following prior work \cite{Deng_2022, ElasticFace, DBLP:conf/cvpr/DengGXZ19}, we monitor model convergence after each epoch using several face verification benchmarks, including LFW \cite{huang:inria-00321923}, CALFW \cite{DBLP:journals/corr/abs-1708-08197}, CPLFW \cite{CPLFWTech}, CFP-FP \cite{c3517bca662f4193a58fd8f9e3145c8f}, and AgeDB-30 \cite{moschoglou2017agedb}.

\subsection{Evaluation Benchmarks} 
Following training, we evaluate the performance of the early exit at 12 depth levels on several widely used FR benchmarks. These include Labeled Faces in the Wild (LFW) \cite{huang:inria-00321923}, Celebrities in Frontal-Profile in the Wild (CFP-FP) \cite{c3517bca662f4193a58fd8f9e3145c8f}, AgeDB30 \cite{moschoglou2017agedb}, Cross-age LFW (CA-LFW) \cite{DBLP:journals/corr/abs-1708-08197}, and CrossPose LFW (CP-LFW) \cite{CPLFWTech}. We report verification accuracies (\%) following the official evaluation protocols for each of these benchmarks. In addition, we evaluated on large-scale evaluation benchmarks, IARPA Janus Benchmark-B (IJB-B) \cite{inproceedingsijbb}, and IARPA Janus Benchmark-C (IJB-C) \cite{DBLP:conf/icb/MazeADKMO0NACG18}. For IJB-C and IJB-B, we used the official 1:1 mixed verification protocol and reported the verification performance as true acceptance rates (TAR) at false acceptance rates (FAR) of $1e-4$. These benchmarks were selected because they are commonly used to evaluate the latest advancements in FR and offer a diverse range of use cases \cite{Deng_2022, wang2018cosfacelargemargincosine, ElasticFace, DBLP:conf/iccv/DanLXD0XS23}. We also evaluate our model on the more challenging TinyFace \cite{DBLP:conf/accv/ChengZG18} benchmark, which consists of unconstrained, low-resolution face images. Through this evaluation, we assess the model’s robustness on a low-quality face dataset, highlighting its ability to generalize beyond controlled scenarios.

\subsection{Exit-Specific Fine-Tuning with Synthetic Data}
To evaluate the effectiveness of the proposed lightweight adaptation strategy, we conduct additional experiments where the final projection layer is fine-tuned for each early exit using a small synthetic dataset. We utilized recent SOTA synthetic data, IDPetrub \cite{boutros2026idperturb}, for finetuning. The dataset contains 0.5M images of 10K identities. This dataset is significantly smaller than MS1MV2 \cite{DBLP:conf/cvpr/DengGXZ19} and is used solely for fine-tuning the final projection layer, enabling efficient adaptation without requiring large-scale labeled data.

\paragraph{Fine-Tuning Setups}
Starting from the pretrained ViT model described in Section \ref{sec:experimental_setup}, we freeze all transformer encoder blocks and fine-tune only the projection layer. For the exit-specific setting, we replace the shared projection function $g(\cdot)$ with exit-dependent layers $\{g_i(\cdot)\}_{i=1}^{L}$, as described in Section~II. Each projection head is optimized independently using the CosFace loss \cite{wang2018cosfacelargemargincosine}. Fine-tuning is performed for 40 epochs using the AdamW optimizer with a learning rate of $1 \times 10^{-3}$ and a batch size of $1024$. To improve generalization, we apply a set of data augmentation techniques, following \cite{DBLP:conf/cvpr/KimS0JL24}.

\paragraph{Evaluation Protocol}
After fine-tuning, each exit-specific model is evaluated using the same benchmarks and protocols described in Section \ref{sec:experimental_setup}. This allows a direct comparison between the original training-free ViT-FREE setup and the fine-tuned variant ViT-FREE$_{FT}$.

\begin{figure}[!t]
  \centering

\begin{minipage}{1\linewidth}
  \centering
    \begin{subfigure}[b]{0.14\linewidth}
      \includegraphics[width=\linewidth, height=1.5cm]{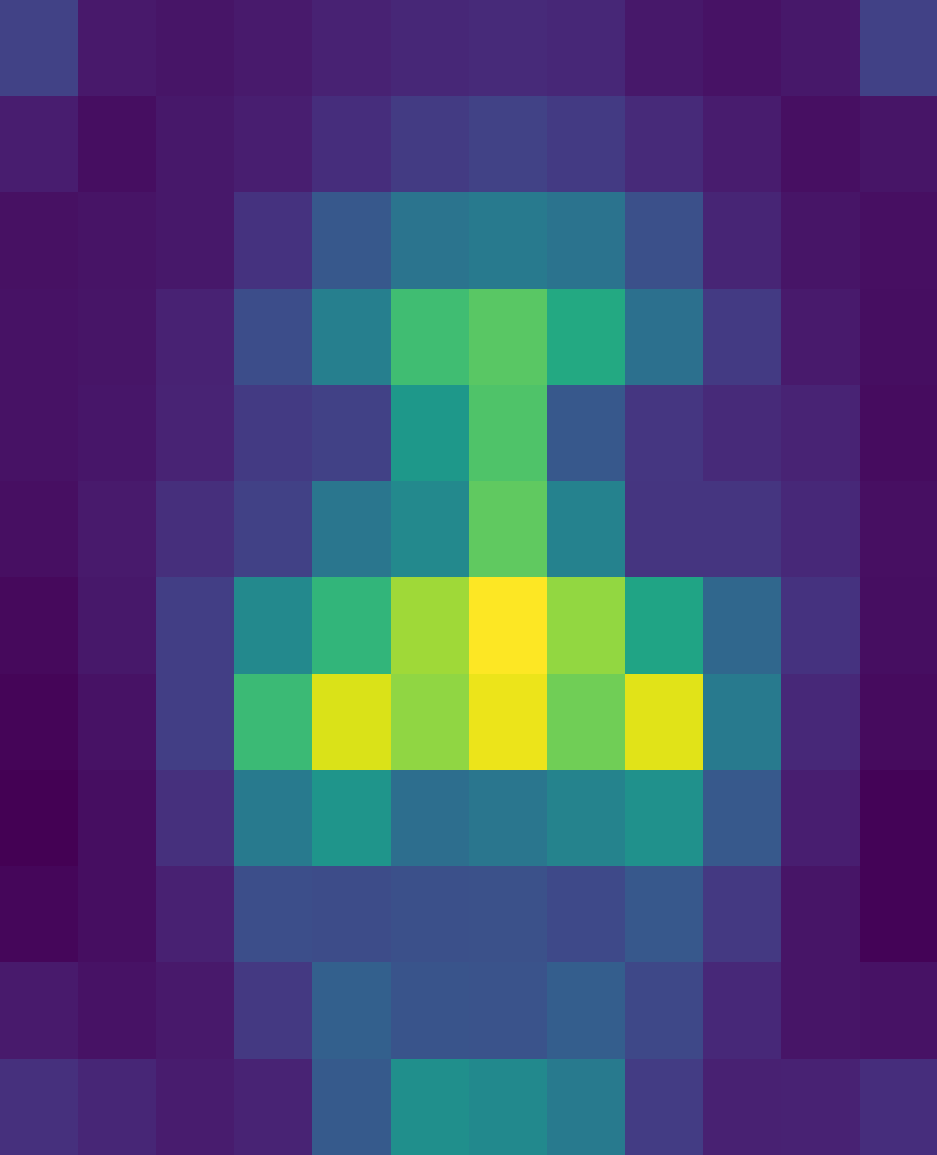}
      \caption{EE1}
  \end{subfigure}
  \hfill
  \begin{subfigure}[b]{0.14\linewidth}
      \includegraphics[width=\linewidth, height=1.5cm]{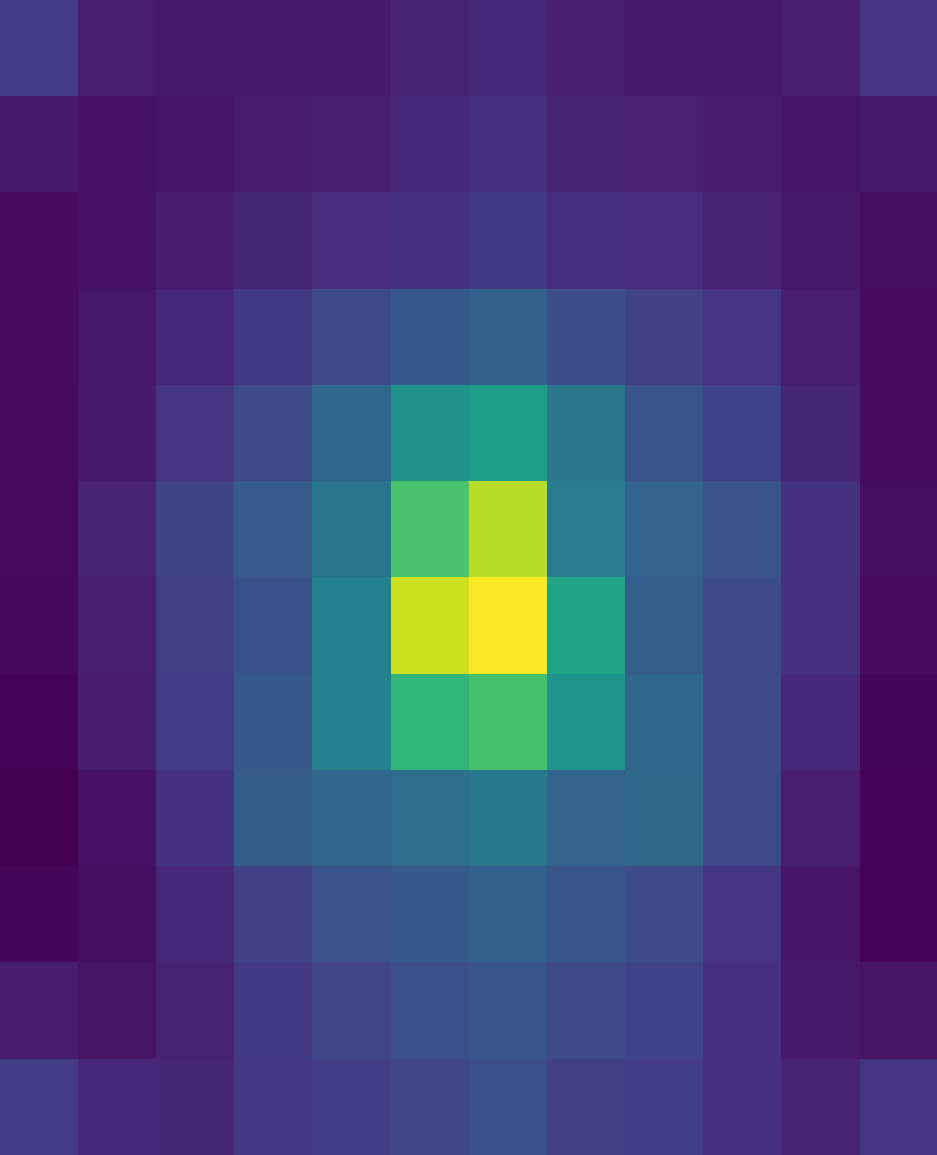}
      \caption{EE2}
  \end{subfigure}
  \hfill
    \begin{subfigure}[b]{0.14\linewidth}
      \includegraphics[width=\linewidth, height=1.5cm]{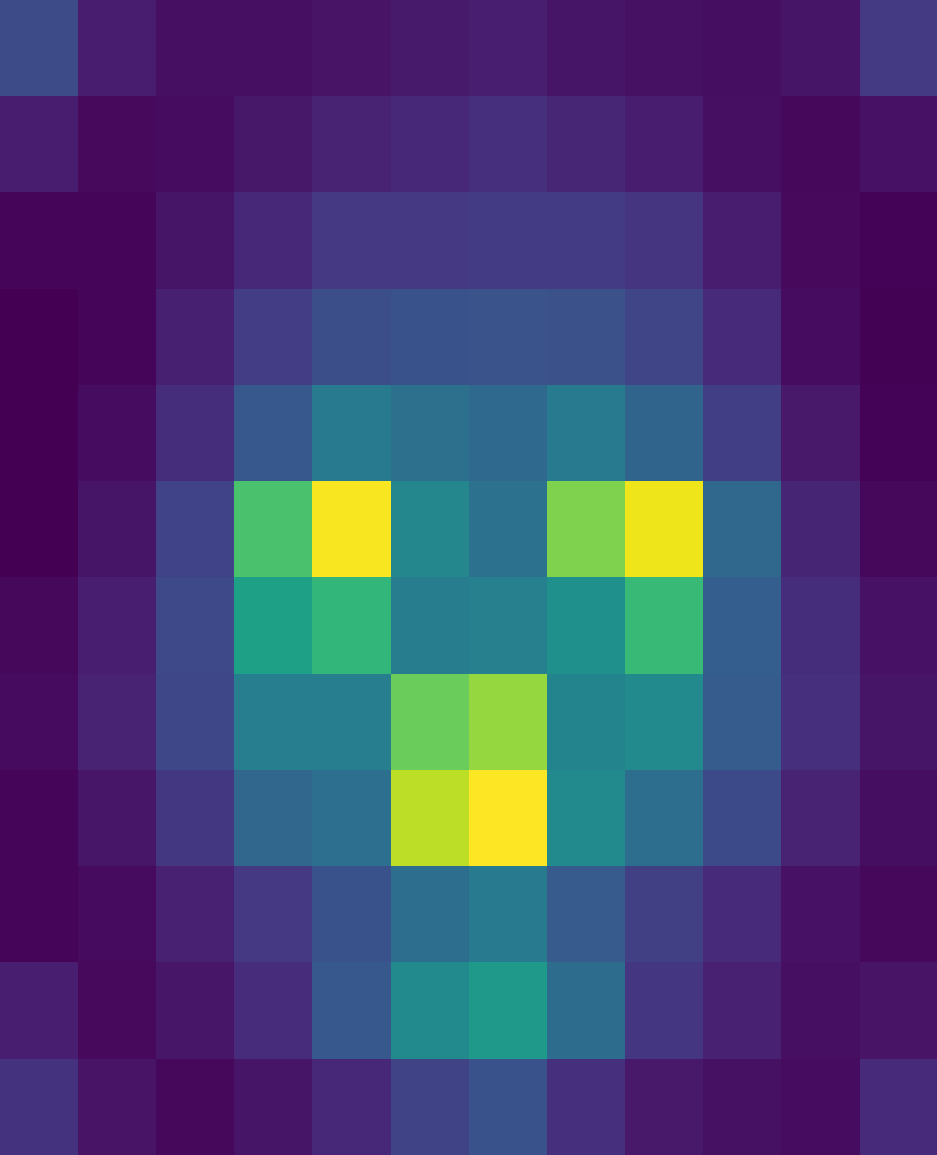}
      \caption{EE3}
  \end{subfigure}
  \hfill
    \begin{subfigure}[b]{0.14\linewidth}
      \includegraphics[width=\linewidth, height=1.5cm]{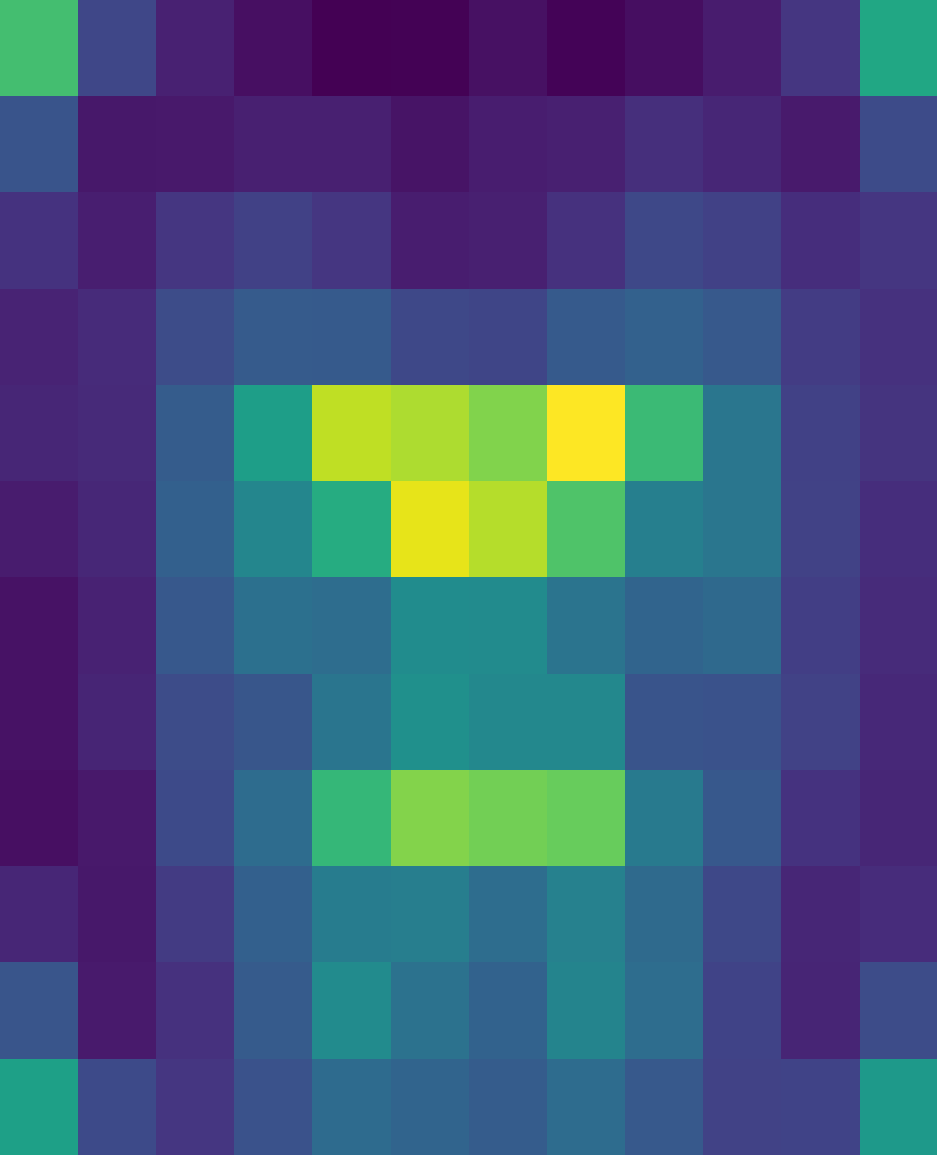}
      \caption{EE4}
  \end{subfigure}
  \hfill
  \begin{subfigure}[b]{0.14\linewidth}
      \includegraphics[width=\linewidth, height=1.5cm]{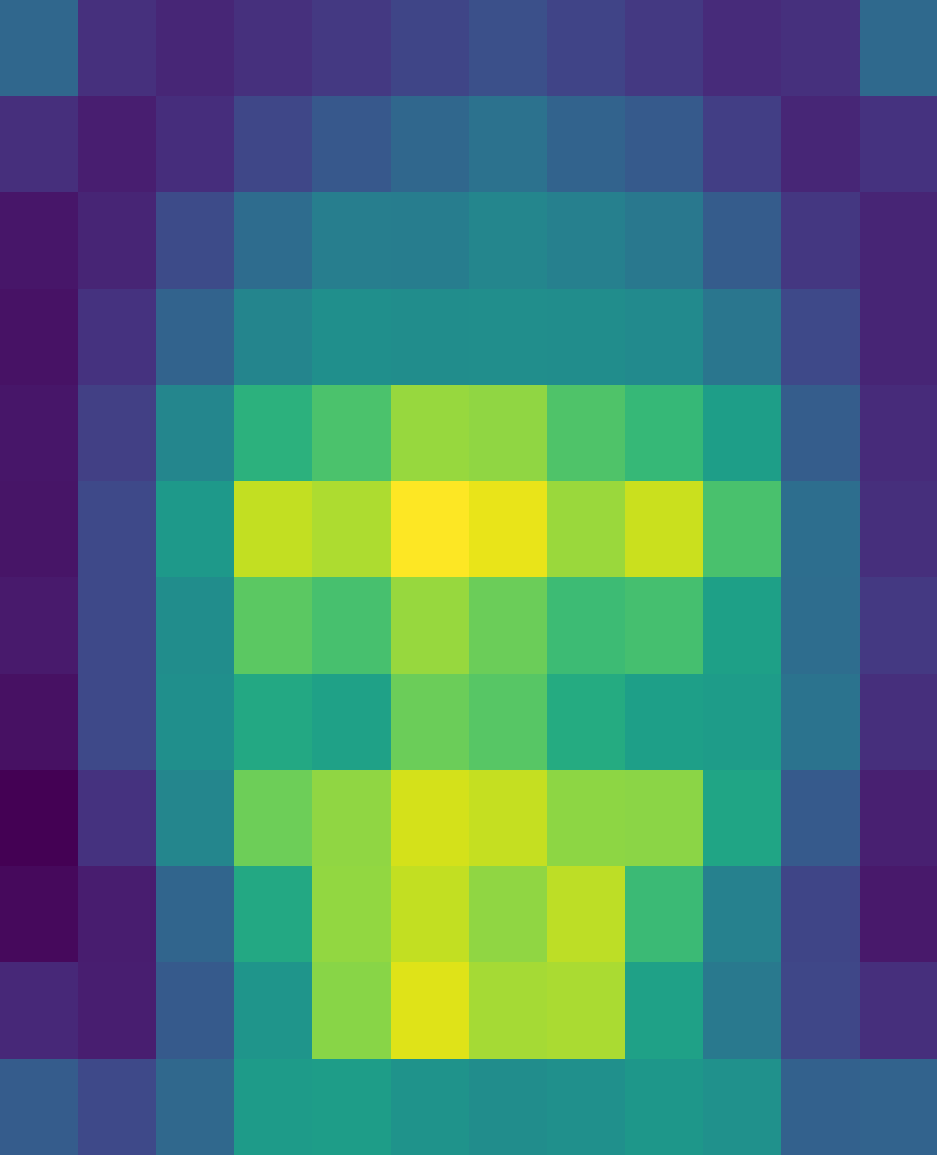}
      \caption{EE5}
  \end{subfigure}
  \hfill
    \begin{subfigure}[b]{0.14\linewidth}
      \includegraphics[width=\linewidth, height=1.5cm]{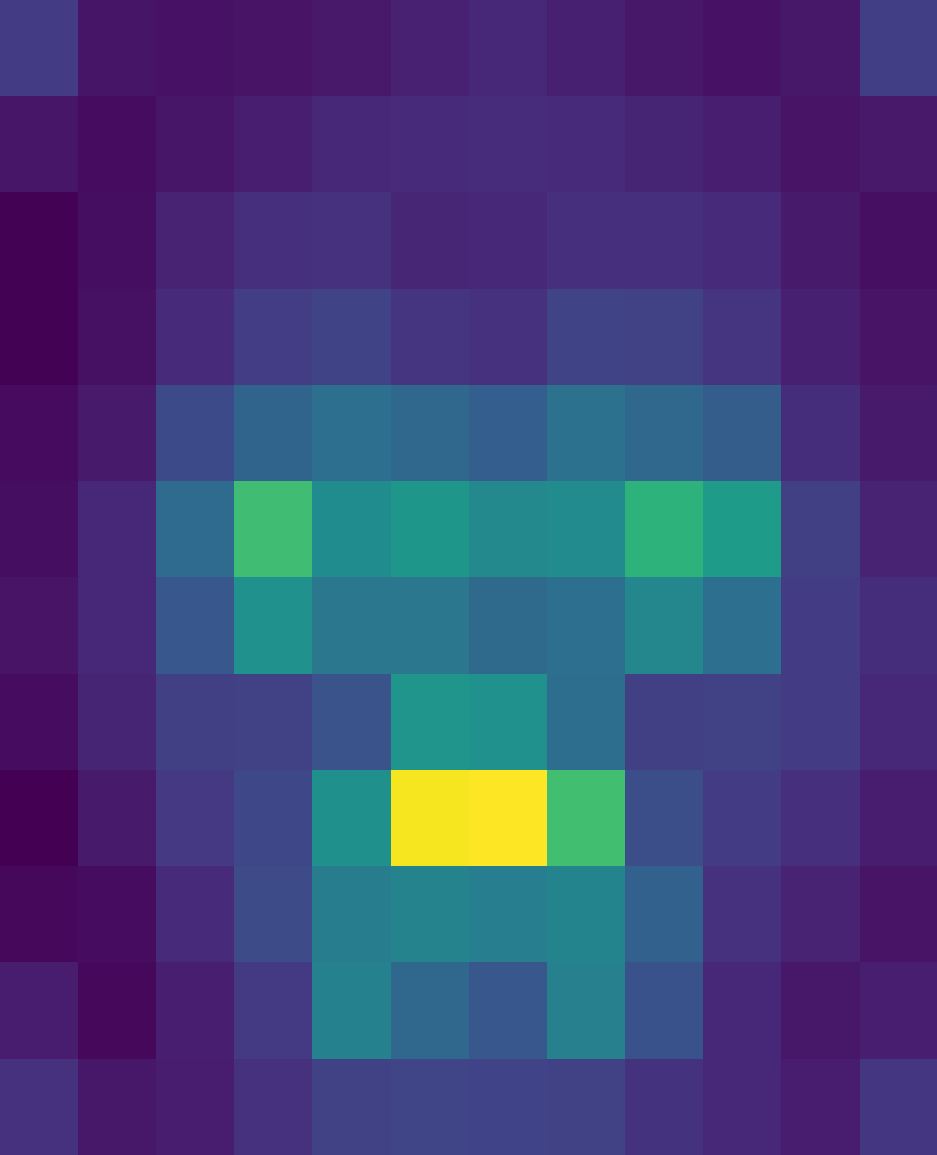}
      \caption{EE6}
  \end{subfigure}
\end{minipage}
  
  \vspace{0.3cm} 

\begin{minipage}{1\linewidth}
  \centering
    \begin{subfigure}[b]{0.14\linewidth}
      \includegraphics[width=\linewidth, height=1.5cm]{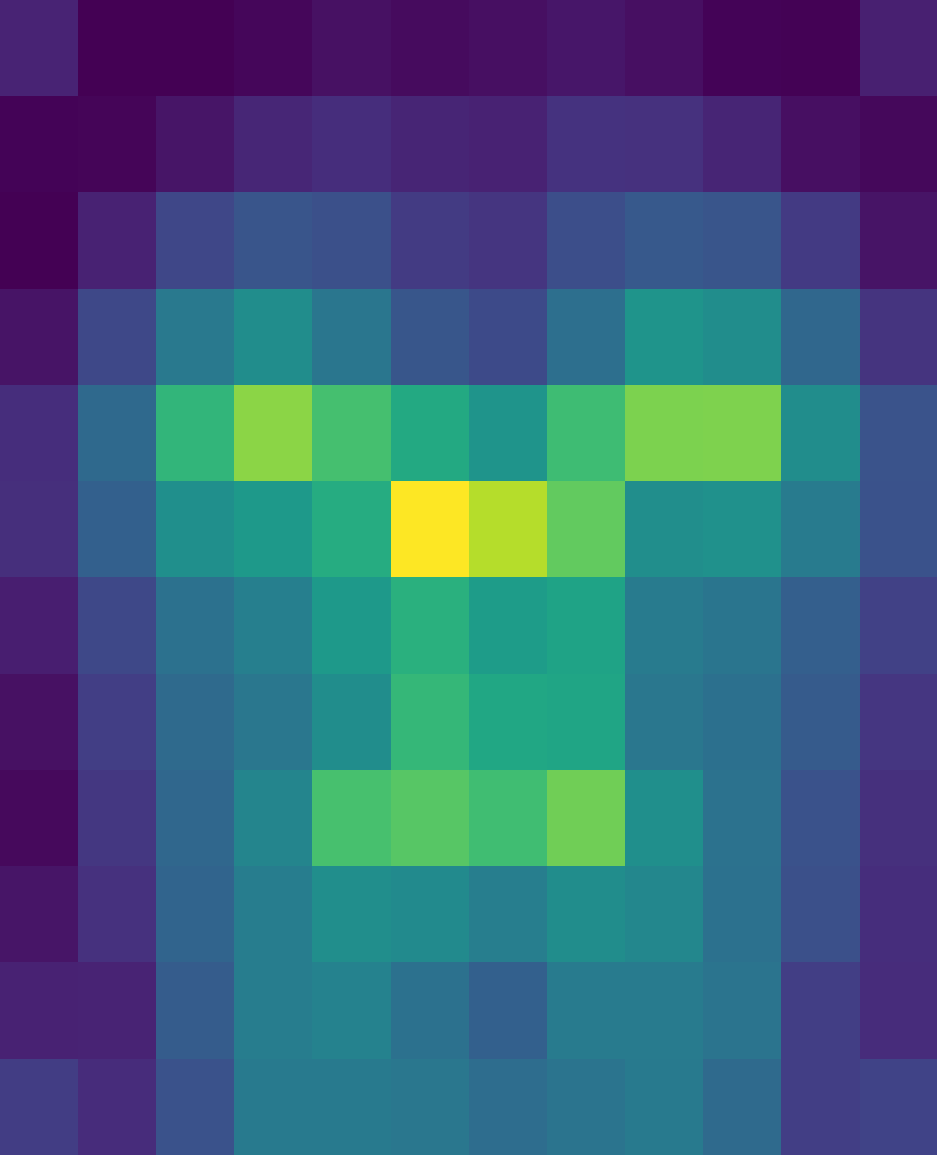}
      \caption{EE7}
  \end{subfigure}
  \hfill
  \begin{subfigure}[b]{0.14\linewidth}
      \includegraphics[width=\linewidth, height=1.5cm]{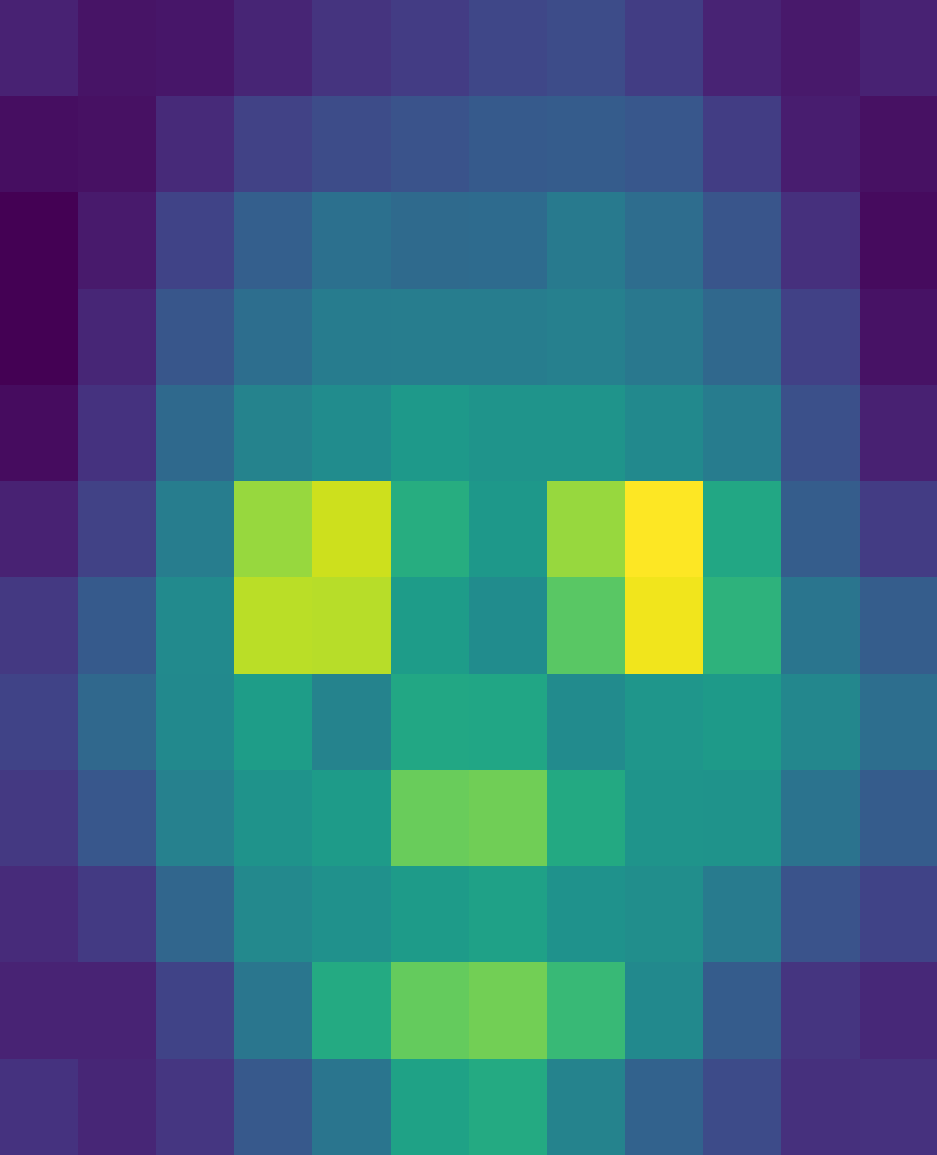}
      \caption{EE8}
  \end{subfigure}
  \hfill
    \begin{subfigure}[b]{0.14\linewidth}
      \includegraphics[width=\linewidth, height=1.5cm]{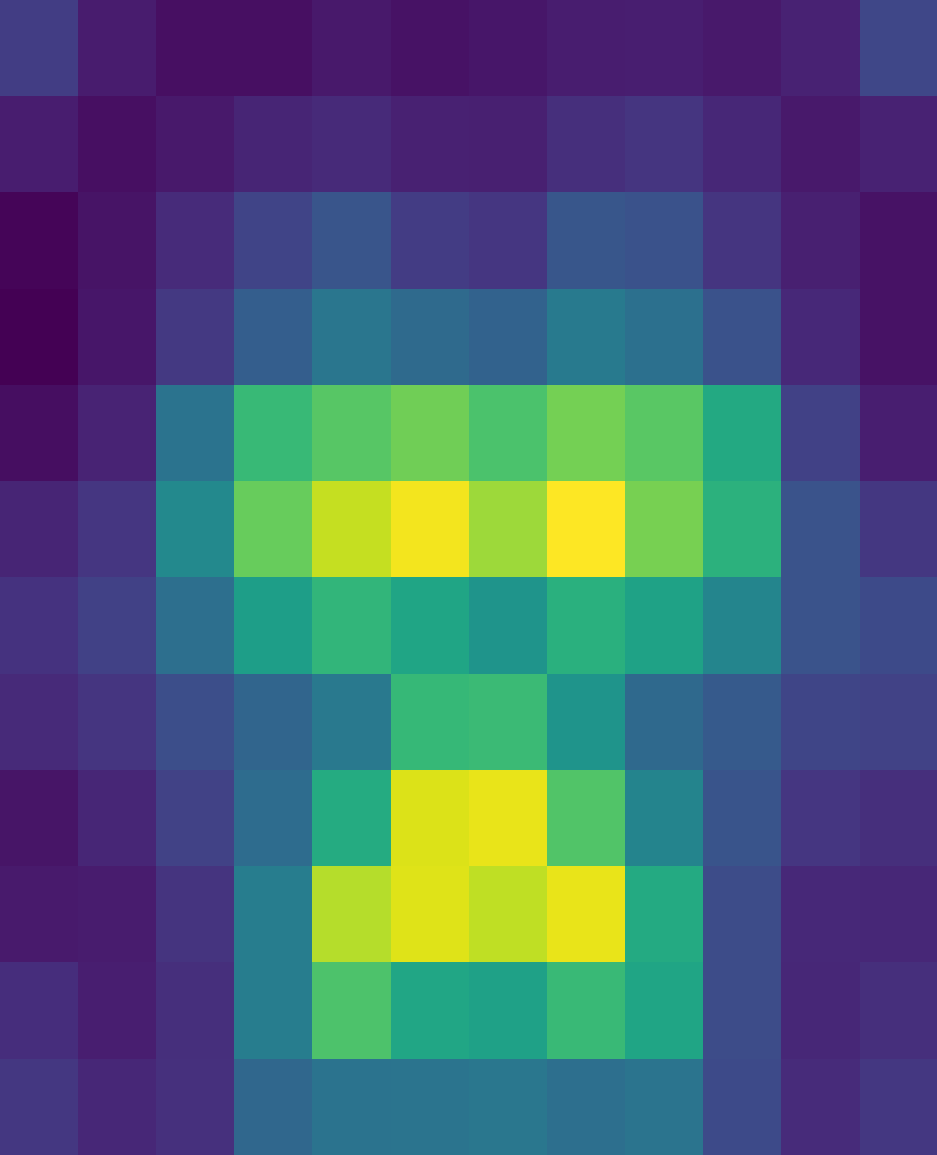}
      \caption{EE9}
  \end{subfigure}
  \hfill
    \begin{subfigure}[b]{0.14\linewidth}
      \includegraphics[width=\linewidth, height=1.5cm]{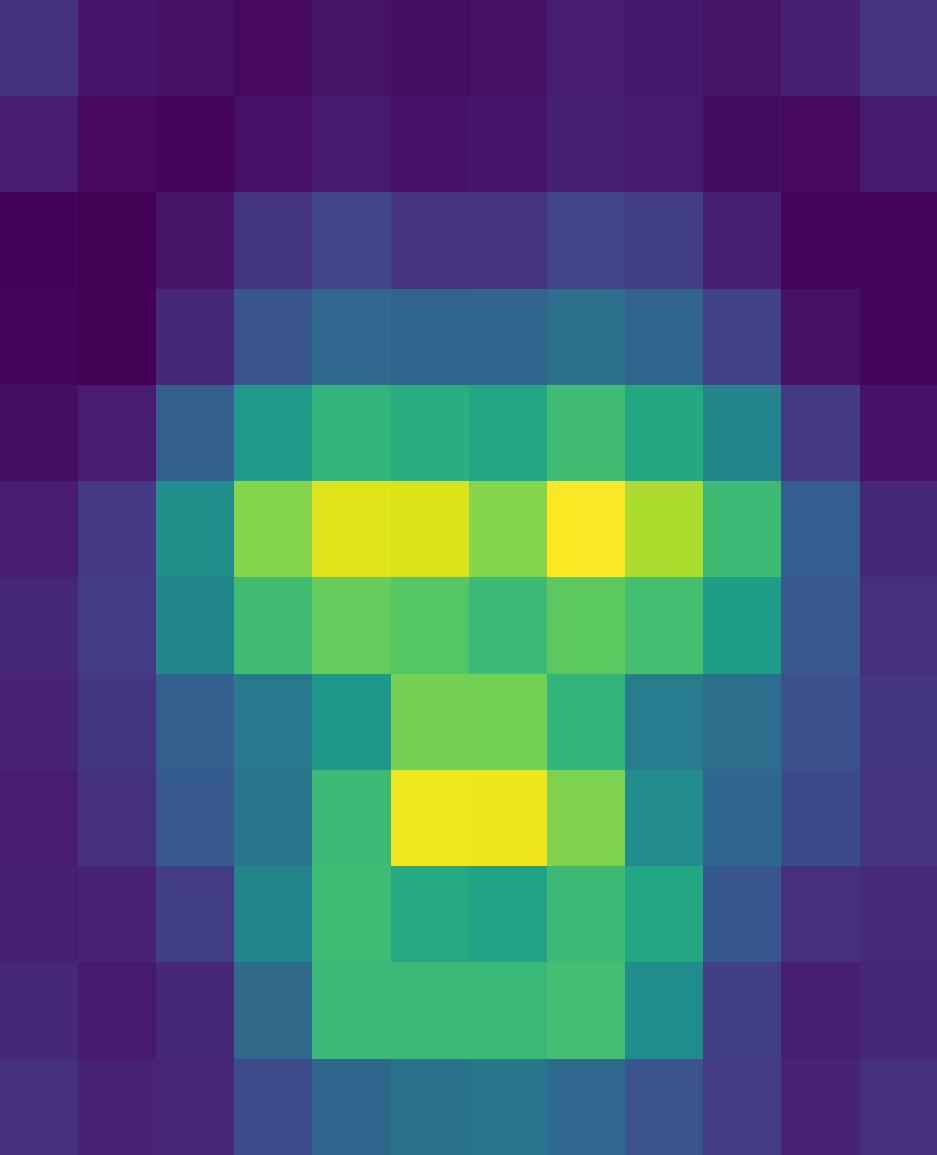}
      \caption{EE10}
  \end{subfigure}
  \hfill
  \begin{subfigure}[b]{0.14\linewidth}
      \includegraphics[width=\linewidth, height=1.5cm]{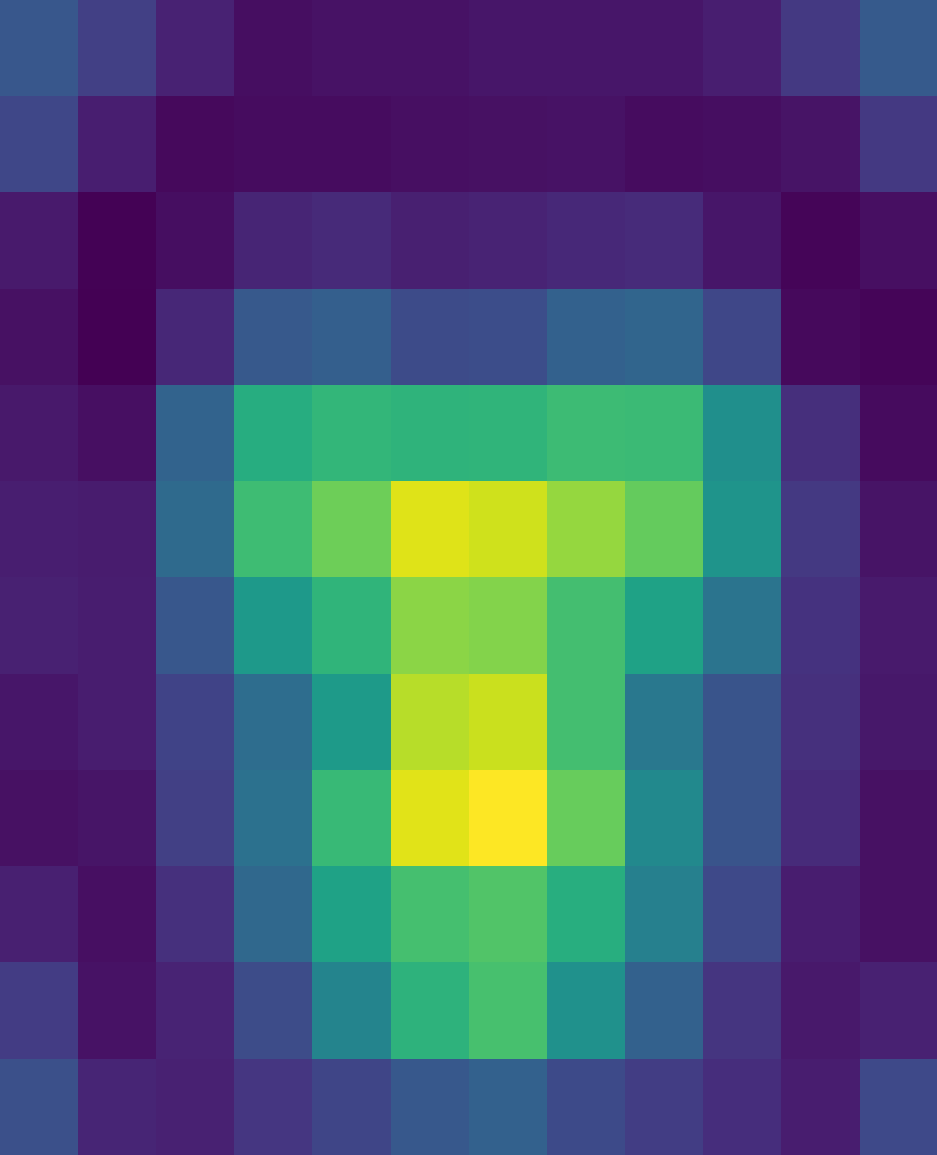}
      \caption{EE11}
  \end{subfigure}
  \hfill
    \begin{subfigure}[b]{0.14\linewidth}
      \includegraphics[width=\linewidth, height=1.5cm]{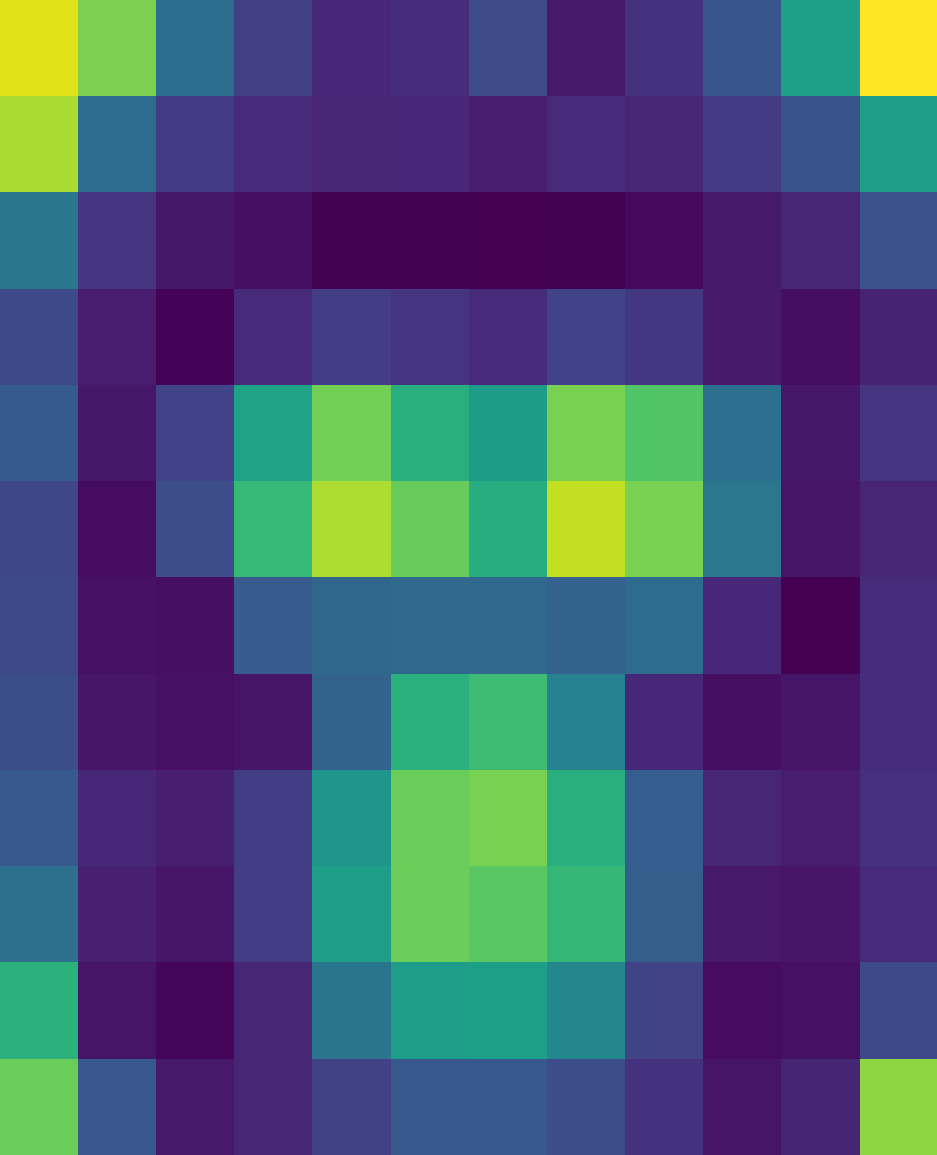}
      \caption{EE12}
  \end{subfigure}
  
\end{minipage}
\caption{Attention maps of ViT-FREE at depths 1–12, corresponding to the attention patterns at intermediate early-exit depths. In the early layers, attention is spread out and mainly concentrated near the center of the image. As the network progresses to deeper layers, attention maps exhibit more defined patterns, with stronger responses around the eyes, nose, and mouth.}
\label{fig:attention}
\vspace{-6mm}
\end{figure}
\vspace{-1mm}
\section{Results} 

\begin{figure*}[ht!]
    \centering

    \begin{subfigure}[b]{0.3\textwidth}
        \centering
        \includegraphics[width=\textwidth]{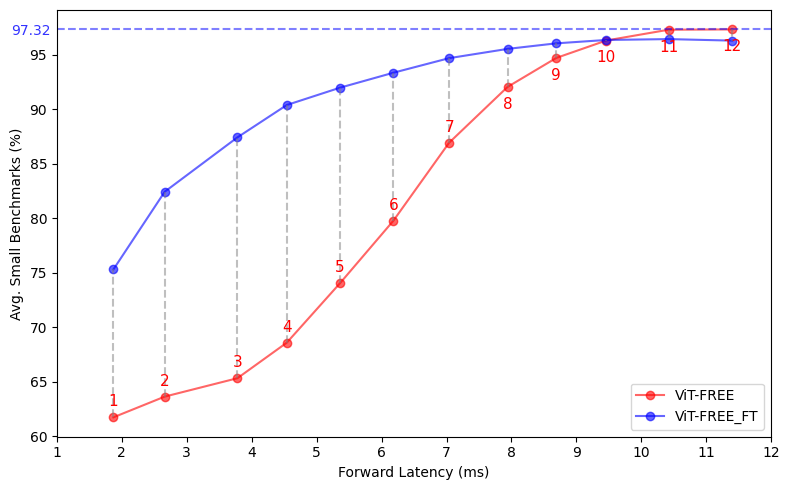}
        \caption{Latency–Accuracy Trade-off on SB}
        \label{fig:img1}
    \end{subfigure}
    \hfill
    \begin{subfigure}[b]{0.3\textwidth}
        \centering
        \includegraphics[width=\textwidth]{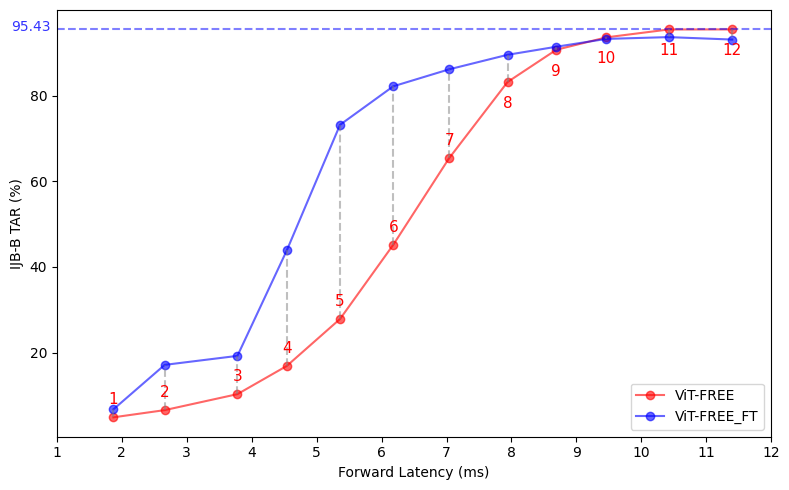}
        \caption{Latency–Accuracy Trade-off on IJBB}
        \label{fig:img2}
    \end{subfigure}
    \hfill
    \begin{subfigure}[b]{0.3\textwidth}
        \centering
        \includegraphics[width=\textwidth]{media/ijbc_latency.png}
        \caption{Latency–Accuracy Trade-off on IJBC}
        \label{fig:img3}
    \end{subfigure}

    \vskip\baselineskip 

    \begin{subfigure}[b]{0.3\textwidth}
        \centering
        \includegraphics[width=\textwidth]{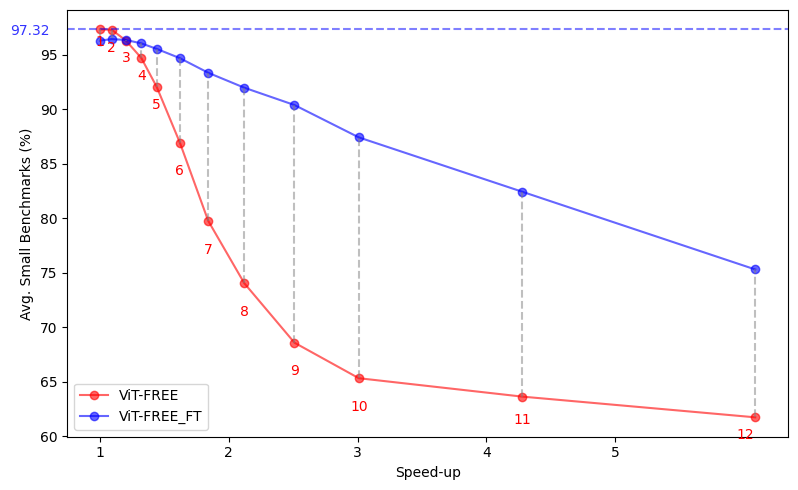}
        \caption{Speedup–Accuracy Trade-off on SB}
        \label{fig:img4}
    \end{subfigure}
    \hfill
    \begin{subfigure}[b]{0.3\textwidth}
        \centering
        \includegraphics[width=\textwidth]{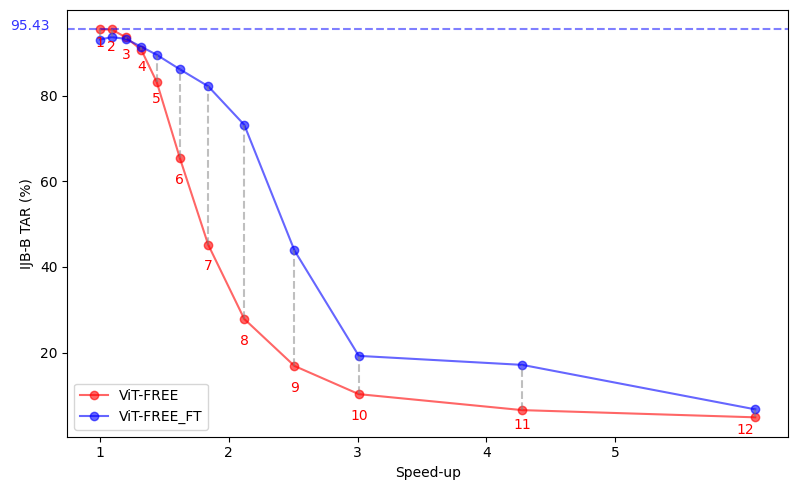}
        \caption{Speedup–Accuracy Trade-off on IJBB}
        \label{fig:img5}
    \end{subfigure}
    \hfill
    \begin{subfigure}[b]{0.3\textwidth}
        \centering
        \includegraphics[width=\textwidth]{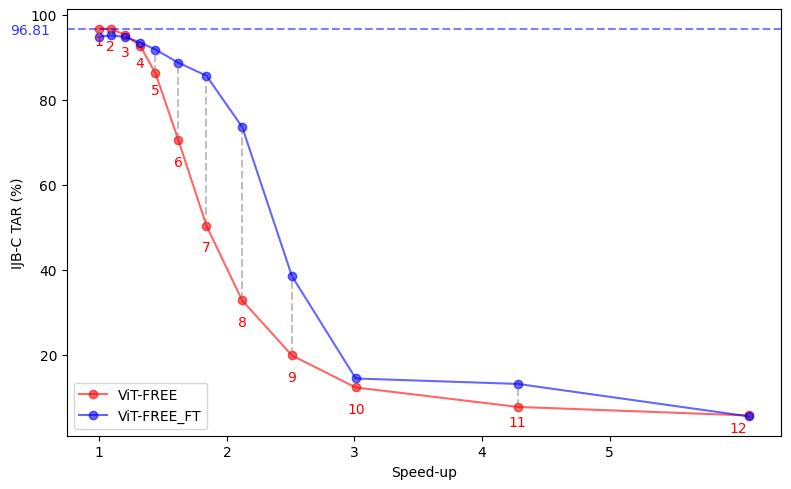}
        \caption{Speedup–Accuracy Trade-off on IJBC}
        \label{fig:img6}
    \end{subfigure}

    \caption{ViT-FREE and ViT-FREE$_{FT}$ efficiency–performance trade-off. SB represents the average over several small benchmarks, as defined in Section \ref{subsec:results}. Each point corresponds to a specific early-exit depth, illustrating the trade-off between computational latency and FR performance. The red curve represents the baseline ViT-FREE, while the blue curve corresponds to the fine-tuned variant ViT-FREE$_{FT}$. Fine-tuning substantially improves the performance of shallow exits, while both models converge to similar performance at deeper layers.}
    \label{fig:plots}
    \vspace{-4mm}
\end{figure*}

\subsection{Accuracy–Efficiency Trade-off for ViT-FREE} \label{subsec:results}
To evaluate the effect of early exiting at different depths on ViT performance for FR, we assessed all 12 exits of ViT-S on the previously presented FR benchmarks summarized in Table \ref{tab:results_table}. In addition to accuracy, we report in Table \ref{tab:results_table_computation} the computational efficiency in terms of FLOPs, forward latency (ms), speed-up, defined as the ratio of the full model’s forward latency to that of the considered early exit, and parameter count. 

Comparing the full model at Exit 12 with Exit 11, the IJB-C \cite{DBLP:conf/icb/MazeADKMO0NACG18} accuracy drops slightly from 96.81 to 96.74, while, surprisingly, benchmarks such as LFW \cite{huang:inria-00321923}, CALFW \cite{DBLP:journals/corr/abs-1708-08197}, and CPLFW \cite{CPLFWTech} show small gains, with TinyFace \cite{DBLP:conf/accv/ChengZG18} experiencing a more noticeable improvement from 68.37 to 68.96. In terms of efficiency, the model at Exit 11 is approximately 9\% faster than the full model. Exiting one layer earlier at Exit 10 leads to a larger drop in IJB-C accuracy, from 96.81 to 95.36, but inference becomes roughly 20\% faster. Beyond Exit 10, the performance decline accelerates considerably, as shown in Figure \ref{fig:ijbc_latency}, suggesting that layers 10–11 offer the optimal trade-off between accuracy and efficiency. To further validate the generalization of our findings, we extend our analysis to a state-of-the-art ViT-based FR model with a larger architecture. Specifically, we evaluate ViT-B KPRPE \cite{DBLP:conf/cvpr/KimS0JL24}, and report the results in Table \ref{tab:results_table_kprpe}. Given the 24-layer depth of ViT-B, results are reported at intervals of two layers (i.e., depths 2, 4, 6, …, 24). The observed behavior follows a similar trend to ViT-S, where performance improves progressively with depth, and later layers provide a more favorable accuracy–efficiency trade-off, confirming the consistency of our findings across architectures.

In addition to the plot illustrating the effect of ViT-FREE early exits at different depths (1–12) on forward latency and IJB-C \cite{DBLP:conf/icb/MazeADKMO0NACG18} performance, we present additional plots for IJB-B \cite{inproceedingsijbb} and SB, where SB denotes the average over LFW \cite{huang:inria-00321923}, CFP-FP \cite{c3517bca662f4193a58fd8f9e3145c8f}, AgeDB30 \cite{moschoglou2017agedb}, CA-LFW \cite{DBLP:journals/corr/abs-1708-08197}, and CP-LFW \cite{CPLFWTech}. We further include the plots for IJB-B, IJB-C, and SB, reformulated to highlight speedup rather than latency. Figure \ref{fig:plots} (a–c) presents the forward latency results, while (d–f) depict the corresponding speedup variants. The x-axis denotes either the model’s forward latency (in milliseconds) or speedup, and the y-axis reports the FR performance on the respective benchmark. Each point represents a specific early-exit depth, thereby illustrating the trade-off between computational efficiency and FR performance. Across the different benchmark plots, we observe that beyond Exit 10, the performance decline accelerates considerably, suggesting that layers 10–11 provide the optimal trade-off between accuracy and efficiency, which further validates our original findings.

\vspace{-1mm}
\subsection{Enhanced Accuracy–Efficiency Trade-off via Fine-Tuned ViT-FREE$_{FT}$} 
\vspace{-1mm}
As described in Section \ref{sec:experimental_setup}, ViT-FREE$_{FT}$ is a fine-tuned variant of ViT-FREE in which the final projection layer of each early exit is independently adapted using a small synthetic dataset, while all transformer encoder blocks remain frozen. To evaluate the effectiveness of this lightweight adaptation strategy, we assess all 12 exits of ViT-FREE$_{FT}$ across the same FR benchmarks and compare them directly against the training-free ViT-FREE baseline. The results are presented in the lower half of Table \ref{tab:results_table}.

Fine-tuning substantially improves the performance of shallow exits, most evidently on the large-scale IJB-C benchmark as shown in Figure \ref{fig:ijbc_latency}. At Exit 7, for instance, ViT-FREE$_{FT}$ achieves an IJB-C TAR of 88.77 compared to 70.49 for the base model. This pattern holds consistently across exits 1 through 9, demonstrating that fine-tuning on synthetic data effectively strengthens early intermediate representations. This can be explained by the disparity in training data scale and quality. The original model was pretrained at full depth on MS1MV2 \cite{DBLP:conf/cvpr/DengGXZ19}, a substantially larger and curated dataset, benefiting from more discriminative feature representations. In contrast, fine-tuning adapts only the projection layer of each individual exit independently on a small synthetic dataset of only 0.5M images \cite{boutros2026idperturb}, which is insufficient to match the representational quality achieved by full-depth pretraining at the deepest exits. In addition to the IJB-C analysis, we present corresponding plots for IJB-B \cite{inproceedingsijbb} and SB, as well as their speedup variants, in Figure \ref{fig:plots}. Across all benchmarks, ViT-FREE$_{FT}$ consistently improves the performance of shallow exits, resulting in a more favorable accuracy–efficiency trade-off at earlier depths compared to the baseline. While both models converge to similar performance at deeper layers, the fine-tuned variant enables stronger performance under tighter computational constraints, shifting the optimal trade-off toward earlier exits.

Additionally, we analyze the cosine similarity between intermediate feature embeddings at each ViT depth (0–10) and the final embedding (depth 11) for both the baseline ViT-FREE and its fine-tuned variant ViT-FREE$_{FT}$, as illustrated in Figure \ref{fig:plot_feature_embedding_similarity}. We observe that the similarity to the final embedding increases progressively across layers (1–10), a trend that holds for both models and is consistent with the feature refinement process described in Section \ref{sec:method}, where each layer incrementally refines the representation and gradually converges toward the final embedding. Notably, the fine-tuned variant (orange bars) consistently exhibits higher similarity to the final embedding at shallow depths (1–9) compared to the baseline ViT-FREE (blue bars), while both models reach comparable similarity levels at deeper layers. This behavior highlights the effectiveness of fine-tuning in accelerating feature refinement throughout the network and further validates our earlier findings that Exit-Specific Fine-Tuning, even with a limited amount of synthetic data, substantially improves the quality of shallow exits (1–9), while providing little to no benefit for deeper exits.

\begin{figure}[!t]
  \centering
   \includegraphics[width=1\linewidth]{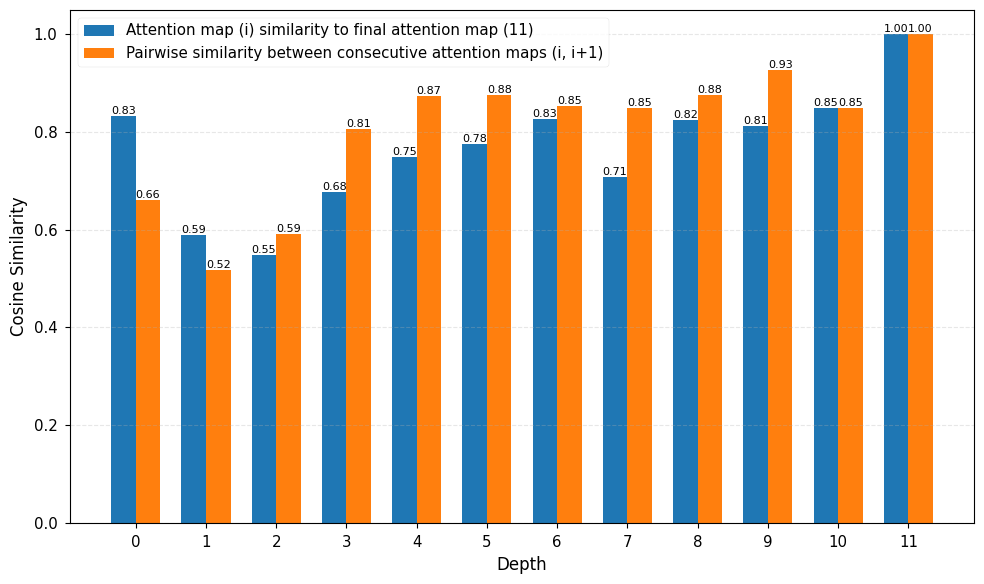}
   \caption{Cosine similarity of Attention maps at different depths (0–11) in a ViT, averaged over the LFW benchmark. Blue bars show similarity between the attention map at depth $i$ and the final attention map, while orange bars show pairwise similarity between consecutive attention maps $(i, i+1)$. Both metrics increase with depth, indicating progressive refinement of attention patterns toward the final layer.}
   \label{fig:plot_attention_map_similarity}
   \vspace{-4mm}
\end{figure}

\subsection{Qualitative Analysis of the Attention Maps}
To better understand the effect of the early exits on the ViT-FREE model, we visualize the attention maps at depths 1–12 in Figure \ref{fig:attention}. These maps were extracted from the last attention layer of each early exit and averaged across $20$ randomly selected identities from the MS1MV2 dataset \cite{DBLP:conf/cvpr/DengGXZ19}, resulting in a total of $1500$ images. In the early layers, attention is diffuse and primarily focused near the center of the image. As we move to deeper layers, attention maps become more structured, highlighting distinct responses around the eyes, nose, and mouth. This indicates that deeper layers prioritize semantically meaningful facial regions over the background. We also observe high attention responses in the corners of the attention maps of the final exit (12). This undesirable effect reduces the interpretability of the attention maps and has been described by Darcet et al. \cite{DBLP:conf/iclr/DarcetOMB24} as artifacts. These artifacts correspond to high-norm tokens appearing in low-informative background regions of the image, which are repurposed by the model for internal computations.

To quantify how attention patterns evolve across depths, we measure cosine similarity between intermediate attention maps, as shown in Figure \ref{fig:plot_attention_map_similarity}. The blue bars indicate the similarity between each intermediate attention map and the final representation at depth 11, while the orange bars capture the similarity between consecutive layers, reflecting the local rate of change. At depth 0, the similarity to the final attention map is relatively high (0.83), but it drops significantly at depths 1 and 2 (0.59 and 0.55), suggesting that the early layers initially diverge from the attention pattern derived directly from the input. From depth 3 onward, the similarity increases across layers. A comparable trend is observed for consecutive-layer similarity, which also rises progressively. The high similarity between consecutive layers in the deeper stages of the network indicates that attention maps undergo only minor changes beyond the midpoint, suggesting that the representations largely stabilize before the final layer.



\section{Conclusion}
In this work, we systematically investigated early exiting in ViTs as an effective and lightweight strategy for efficient FR. By exploiting the uniform feature dimensionality across transformer encoder blocks, we introduced \textbf{ViT-FREE}, a simple yet effective framework that enables face verification directly from intermediate representations without modifying or retraining the backbone model. Through extensive experiments across multiple benchmarks, we demonstrated a clear and consistent accuracy-efficiency trade-off across exit depths. In particular, later exits (e.g., layers 10--11) provide an optimal balance, achieving up to 20\% inference speedup with only marginal performance degradation on challenging large-scale benchmarks such as IJB-C, while earlier exits exhibit a more pronounced drop in performance. 
To further enhance early-exit performance, we proposed \textbf{ViT-FREE$_{FT}$}, a lightweight exit-specific fine-tuning strategy that adapts only the projection heads using a small synthetic dataset. This approach significantly improves shallow exits while preserving efficiency and leaving deeper representations largely unaffected.
Overall, our findings highlight that intermediate ViT representations are sufficiently discriminative for FR, enabling efficient inference without full-depth computation. This provides practical insights for deploying ViT-based FR systems under strict latency and resource constraints.


\section{Ethical Impact Statement}
Our research on early exiting in ViT-based FR enhances computational efficiency, making accurate recognition systems more accessible for deployment on resource-constrained devices and enabling energy-efficient, real-time applications. This can support beneficial use cases such as on-device processing, reduced latency, and improved privacy through local inference. While early exiting introduces an inherent accuracy–efficiency trade-off, our work emphasizes the importance of carefully selecting exit points that balance performance and resource usage. We highlight that such systems should undergo thorough evaluation to ensure reliability, particularly in sensitive or high-stakes applications. We advocate for responsible deployment within established legal and regulatory frameworks (e.g., GDPR), accompanied by safeguards such as transparency, user consent, and, where appropriate, human oversight. By promoting efficient yet accountable FR systems, this work contributes to the development of more sustainable and privacy-conscious AI technologies.

\clearpage



{\small
\bibliographystyle{ieee}
\bibliography{egbib}
}

\end{document}